\documentclass[11pt]{article}

\usepackage[preprint]{acl}

\usepackage{times}
\usepackage{latexsym}
\usepackage{url}
\usepackage[T1]{fontenc}
\usepackage[utf8]{inputenc}

\usepackage{microtype}

\usepackage{inconsolata}

\usepackage{graphicx}

\usepackage{booktabs}
\usepackage{amsmath}
\usepackage{amssymb}
\usepackage{xcolor}
\usepackage{fvextra}
\usepackage{makecell}
\usepackage{paralist}
\expandafter\def\expandafter\UrlBreaks\expandafter{\UrlBreaks\do\-}
\DeclareUnicodeCharacter{200E}{}
\DeclareUnicodeCharacter{202C}{}
\newcommand{\method}{\textsc{DeRef}}

\title{Judges Follow the Reference: The Derivation Gap in Verifying Tool-Call Arguments}

\author{
 \textbf{Linhai Ma\textsuperscript{1,2}},
 \textbf{Yan Wang\textsuperscript{1,2,*}},
 \textbf{Lingfei Qian\textsuperscript{1,2,*}},\\
 \textbf{Yupeng Cao\textsuperscript{1,2}},
 \textbf{Víctor Gutiérrez-Basulto\textsuperscript{3}}
\\
\\
 \textsuperscript{1}The Fin AI, USA,
 \textsuperscript{2}Yale University, USA,
 \textsuperscript{3}Cardiff University, UK
\\
 \small{
   \textbf{Correspondence:} {\{wy2266336,lfqian94\}@gmail.com}
 }
}

\begin{document}
\maketitle

\begin{abstract}
Tool-call judges evaluate argument values proposed by LLM agents.
We find that they accept a correct value less often when their input holds no reference for it, that is, no text that says what the value should be.
This happens when the value is a default value from the function schema that the user did not mention, when it has to be inferred, or when the text it comes from was cut from the judge input.
We call this drop in acceptance the \emph{derivation gap}.
It appears across all seven judges we evaluate, including System One models CLM and Jev.
The gap is not only a truncation artifact: correct default values have the largest gap on a benchmark with no truncation.
A controlled source-relocation test shows that judges follow the reference in their input: they favor a visible incorrect value over a correct value whose source is hidden.
We introduce \method{} (Derived Reference): a separate model reads the full history without seeing the candidate and writes a reference for the judge.
Across three benchmarks, \method{} improves overall argument AUROC for all judges tested.
On the controlled test, it raises AUROC from chance or below to between 98.40 and 99.59.
The code is available at \url{https://github.com/SarielMa/DeRef_release}
\end{abstract}

\section{Introduction}
\begin{figure*}[t]
  \centering
  \includegraphics[width=\textwidth]{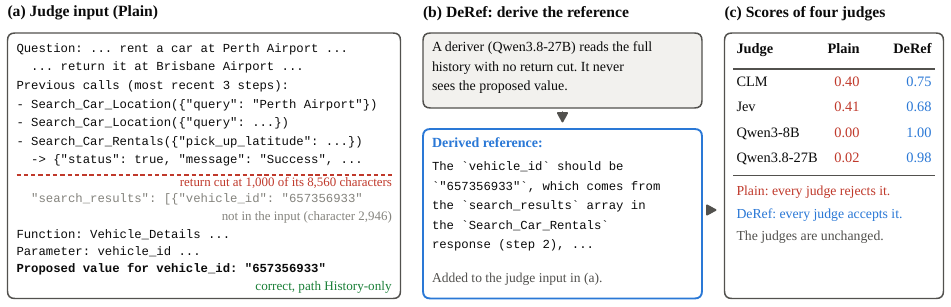}
  \caption{A correct argument on the \textsc{History-only} path, from CFB. (a) The judge input keeps the first 1,000 characters of the search result, and the correct vehicle ID appears later. (b) \method{} asks a deriver to read the full history without the candidate and to write the expected value and its source. The reference goes into the same judge input. (c) Scores of the four judges, rounded to two decimals. With Plain input, every judge scores the value below 0.5 and below its own acceptance threshold $\tau$ (Section~\ref{sec:gap}); for Qwen3-8B, whose $\tau$ on CFB is 0.002, the unrounded score is below that. With \method{}, every judge scores it above both.}
  \label{fig:overview}
\end{figure*}

LLM agents act through tool calls.
% When real models call tools, a wrong argument value is their most common error in FC-RewardBench (FCRB) \citep{agarwal2025toolrm}.
Wrong argument values are among the most common errors in the tool calls of LLM agents, as in FC-RewardBench (FCRB) \citep{agarwal2025toolrm}.
% In FC-RewardBench (FCRB), a wrong argument value is the most common error in tool calls made by real models \citep{agarwal2025toolrm}.
% Agent pipelines therefore use judges to check tool calls, as reward models \citep{agarwal2025toolrm,li2026toolrm}, as process reward models \citep{lin2026toolprm}, and as guards before execution \citep{she2026safeguarding}.
Agent pipelines therefore use judges to check tool calls, as reward models \citep{agarwal2025toolrm,li2026toolrm,lin2026toolprm} and guards before execution \citep{she2026safeguarding}.
% Many checks work argument by argument, such as BFCL, ProvenanceGuard, and ToolPRM \citep{patil2025berkeley,she2026safeguarding,lin2026toolprm}.
Many checks operate at the argument level, from the BFCL benchmark to ProvenanceGuard and ToolPRM \citep{patil2025berkeley,she2026safeguarding,lin2026toolprm}.

An argument judge sees the context and a proposed value, the \emph{candidate}, and decides whether it is right.
But a correct value can reach a call in several ways.
It can be copied from visible text, taken from an earlier tool return, filled in from a default that the function schema specifies when the user does not mention the argument, reformatted, or inferred.
We call each way a \emph{derivation path}.
Figure~\ref{fig:overview} shows one such path: a correct vehicle ID taken from an earlier tool return.
The source of the ID is not visible to the judge, and all four judges reject the value.
Correct arguments that the input does not show directly, like this ID, are accepted much less often than those it does.
We call this drop in acceptance rate the \emph{derivation gap}.

% An argument judge sees the context and a proposed value, the \emph{candidate}, and decides whether it is right.
% But a correct value can reach a call in several ways.
% It can be copied from visible text, taken from an earlier tool return, filled in from a default that the function schema specifies when the user does not mention the argument, reformatted, or inferred.
% We call each way a \emph{derivation path}.
% % Figure~\ref{fig:overview} shows a correct vehicle ID copied from an earlier search result.
% % The judge input keeps only the first 1,000 characters of that result, and the ID comes later.
% % All four judges in the Figure~\ref{fig:overview} reject the value.
% Figure~\ref{fig:overview} shows one such path: a correct vehicle ID taken from an earlier tool return.
% The source of the ID is not visible to the judge, and all four judges reject the value.

% % We measure how often judges accept correct arguments on each path.
% % Correct arguments that the input does not show directly, such as the vehicle ID in Figure~\ref{fig:overview}, are accepted much less often.
% % We call this drop in acceptance rate the \emph{derivation gap}.
% We measure how often judges accept correct arguments on each path.
% Correct arguments that the input does not show directly, such as the vehicle ID in Figure~\ref{fig:overview}, are accepted much less often than those it does.
% We call this drop in acceptance rate the \emph{derivation gap}.
The gap appears in all seven judges we study.
Two are System One judges, scoring models that return a probability rather than text \citep{kwok2026contrastive,almeida2026introducing}.
Five are LLM judges from the Qwen, Llama, and Mistral families \citep{yang2025qwen3,qwenteam2026qwen3827b,grattafiori2024llama3,jiang2023mistral}.
Scaling the judge from 4B to 27B parameters does not remove the gap.
The gap is also not only a matter of missing information: on FCRB, where nothing is cut and each default is shown, correct defaults have the largest gap.

Controlled tests point to one cause: a judge tends to accept a value that is stated in its input, and to reject one that is not.
It does not reliably work out the correct value on its own.
When we hide the source of a correct value and leave a wrong value visible, AUROC falls to chance or below for every judge tested.
Giving the judge the full history narrows the gap on values from earlier steps.
It does not help on values that match no span of the history.
But when the input states the expected value and the reason for it, judges accept values they rejected before.

This points to a simple lever.
If judges rely on a reference in their input, we can supply one.
\method{} (Derived Reference) asks a model to read the full history without seeing the candidate.
The model writes down the expected value of each argument and where it comes from (Figure~\ref{fig:overview}b).
The judge then scores the candidate with this reference in its input.
In Figure~\ref{fig:overview}c, all four judges then accept the vehicle ID.
Unlike judges that write their own answer first, \method{} uses a separate model that reads the full history.

\smallskip \noindent 
\textbf{Our contributions are as follows:}
\textbf{(1)} We identify a systematic derivation gap across seven argument judges, from System One models to a 27B LLM.
On ComplexFuncBench (CFB) and FinTrace it lies between 22.67 and 36.13 points for every judge, and on FCRB, with no truncation, correct defaults have a gap between 14.40 and 50.51 points.
\textbf{(2)} Through controlled source relocation and reference substitution, we show that judgments track explicit reference values in the input, even when these favor an incorrect candidate over the correct one.
\textbf{(3)} We introduce \method{}, which derives a shared reference from the full history without seeing the candidate.
Without retraining or modifying the judges, it improves overall argument AUROC for every judge tested, on all three benchmarks.
Averaged over tested judges, it is significantly above the strongest prior method, RevisEval, on every benchmark.
On the controlled test with a hidden source, it lifts every judge from chance or below to between 98.40 and 99.59.

% ----------------------------------------------------------------------------
% Abstract and Section 1 (Introduction)
% How Victor's comments were addressed
% (1) "What is a schema default?": the abstract no longer uses the bare term.
%     The finding is now stated in two sentences. The first gives the main
%     result. The second lists the three cases and describes the default in
%     plain words (a default value from the tool schema). The introduction describes it the same way.
% (2) "Exemplify using Figure 1": the sentence on correct arguments that the
%     input does not show directly now points to the vehicle ID in Figure 1.
% (3) "I would start a new sentence here": the sentence on the seven judges
%     is split into four.
% (4) "What is FCRB?": the abbreviation is now introduced in the first
%     paragraph, where FC-RewardBench is first mentioned.
% (5) "Didn't we have 6?": the introduction now gives only one judge count,
%     the seven judges of the gap study. The other places say "every judge
%     tested". The judge sets are stated once, in Section 3.2.
% (6) "Exemplify with Figure 1b": the description of DeRef now points to
%     Figure 1b, and one sentence gives the outcome in Figure 1c.
% (7) "Incredibly long and difficult sentence": contribution (3) is split
%     into three sentences.

\section{Related Work}
\label{sec:related}

\smallskip  \noindent \textbf{Judges for tool-calling.}
Benchmarks check calls against gold answers, per argument \citep{patil2025berkeley} or per call \citep{zhong2025complexfuncbench}; without gold answers, learned judges score whole calls \citep{agarwal2025toolrm,li2026toolrm} or each argument as a step \citep{lin2026toolprm}, and benchmarks of such judges find that they tend to label steps as correct and agree less with a reference as tasks get harder \citep{li2026toolprmbench,fan2026agentprocessbench,verma2026agentjudgebench}.
Guards check calls before they run: ProvenanceGuard finds evidence for each argument with the value in hand \citep{she2026safeguarding}, PACT and Agent-Sentry track where each value came from \citep{fan2026granularity,sequeira2026agentsentry}, and TwinCheck repairs a flagged call from the trace \citep{dai2026twincheck}.
Jev trails GPT-6 where the verdict has to be derived rather than read off the text \citep{li2026jevjudge}, and factuality metrics drop on summaries that need deeper reasoning \citep{ramprasad2025automatic}.
To our knowledge, none of these studies splits judge errors by derivation path.

\smallskip  \noindent \textbf{Reference-guided judging.}
Judges grade better with a reference, and several have the judge write its own answer before seeing the candidate \citep{zheng2023judging,zeng2024llmbar,li2026baer,zhou2026convincing,andrade2026let,li2026march,dhuliawala2024chain,raghavendra2026agentic}, or revise the candidate into one \citep{zhang2025reviseval}; without a reference, judges agree less with experts and credit wrong answers \citep{krumdick2025free,chalamalasetti2026llm}, and with a wrong reference they reject answers that match it \citep{lee2026judging}.
These judges write the reference from the question and their own knowledge, which is often not enough for an argument judge, whose expected value can sit in a tool return cut from its input, and a reference the judge writes from its own input, as in LLMBar, stays below \method{} (Section~\ref{sec:main}).

\smallskip  \noindent \textbf{Correct content without support.}
A value without support in the context is not always wrong: unsupported entities in summaries are often factual \citep{cao2022hallucinated}, and RAG models often answer correctly when the context is insufficient \citep{joren2025sufficient}.
Tool-call arguments make this common, since a correct default, format change, or inferred value has no matching span in the dialogue, and we measure how often judges treat such values as errors.
A source in the context can also go unused \citep{modarressi2025nolima,kate2025longfunceval}.
App.~\ref{app:related} discusses each line of work in more detail.

\section{The Derivation Gap}
\label{sec:problem}
% \nb{Provide a short paragraph explaining what is going to happen in this section, which is quite long. In particular, how are 3.1, 3.2 and 3.3 related}
This section identifies the derivation gap and investigates its causes.
Section~\ref{sec:setting} defines the judging task and five derivation paths based on how argument values relate to the history and the function schema.
Section~\ref{sec:gap-setup} describes the benchmarks, judges, and judge input.
Section~\ref{sec:gap} measures differences in acceptance rates for correct values across paths.
Finally, Section~\ref{sec:follow} uses controlled tests to examine the roles of missing information and explicit reference values, motivating the introduction of \method{} in Section~\ref{sec:method}.

\subsection{Argument-Level Judging and Derivation Paths}
\label{sec:setting}

\smallskip  \noindent \textbf{Task.}
An agent acts in a state with history $H$, which contains the user turns, earlier tool calls, and their full returns.
The agent proposes a call to a function $f$ with $k$ arguments, where argument $i$ has a name $a_i$ and a value $v_i$.
An argument judge checks one value at a time using a context $X$ built from $H$ and the schema of $f$.
$X$ does not need to contain all of $H$ (Section~\ref{sec:gap-setup}), so the text that determines a correct value can be present in $H$ but absent from $X$.
For each argument, the judge returns a score $s(v_i \mid a_i, X) \in [0,1]$, representing its estimated probability that $v_i$ is correct.

\smallskip  \noindent \textbf{Derivation paths.}
A correct value is not always a copy of text that the judge can see.
The derivation of a value $v$ can be described by a \emph{source} $e$ and a \emph{transformation} $t$.
The source is a span of $H$ or a field of the function schema, such as a default.
The transformation copies the source, changes its format, or infers a new value from it, so that $t(e) = v$.
Each candidate value, whether right or wrong, has a \emph{derivation path}, determined by text-matching rules (Appendix~\ref{app:census}).
The paths capture how directly a value can be checked against the dialogue shown to the judge.
There are five path types, and a value takes the first one that matches:
\begin{itemize}
\setlength{\itemsep}{0pt}
\setlength{\parsep}{0pt}
\setlength{\parskip}{0pt}
  \item \textsc{Visible}: $v$ appears in the dialogue part of $X$, that is, the user turns, shown calls, and retained portions of their returns.
  \item \textsc{History-only}: $v$ appears in $H$ but not in the dialogue part of $X$. The matching text was cut or dropped, as for the vehicle ID in Figure~\ref{fig:overview}.
  \item \textsc{Format}: $v$ matches a span of $H$ after a format-normalization step, such as converting two date expressions to the same form.
  \item \textsc{Default}: $v$ matches a schema default or a ``don't care'' value explicitly specified in the parameter description.
  \item \textsc{No-source}: no match is found under the preceding rules.
\end{itemize}
Correctness is determined by the benchmark's gold call, not by the assigned path.

\subsection{Benchmarks and Judges}
\label{sec:gap-setup}

\smallskip  \noindent \textbf{Benchmarks.}
We use three tool-calling benchmarks, each with a gold call at every step (Appendix~\ref{app:data}).
FC-RewardBench (FCRB) gives 630 single-turn requests, each with a correct call and a wrong call made by a real model \citep{agarwal2025toolrm}.
ComplexFuncBench (CFB) has multi-step tasks with long tool returns \citep{zhong2025complexfuncbench}.
Its wrong calls come from the trajectories of 30 models \citep{agentsuite2026complexfuncbench}, which gives 347 correct and 644 wrong calls.
FinTrace gives 2,334 golden calls for financial queries \citep{cao2026fintrace} and 2,334 wrong calls, each made from a golden call by changing one argument by rule.

\smallskip  \noindent \textbf{Judges.}
We study seven judges (Appendix~\ref{app:impl}).
CLM \citep{kwok2026contrastive} and Jev \citep{almeida2026introducing} are System One judges, scoring models that return a probability for a question about their input.
The other five are LLM judges: Qwen3 at 4B and 8B \citep{yang2025qwen3}, Qwen3.8-27B \citep{qwenteam2026qwen3827b}, Llama-3.1-8B-Instruct \citep{grattafiori2024llama3}, and Mistral-7B-Instruct-v0.3 \citep{jiang2023mistral}.
Their score is the probability of ``Yes'' at the first output token, with thinking turned off.
Section~\ref{sec:gap} reports all seven judges.
The controlled tests of Section~\ref{sec:follow} and the main experiments of Section~\ref{sec:exp} use four of them: CLM, Jev, Qwen3-8B, and Qwen3.8-27B.

\smallskip  \noindent \textbf{Judge input.}
All judges see the same context $X$, which we call the \emph{Plain input}.
We choose a truncated input on purpose: it keeps the last few steps and cuts each return to 1,000 characters (Appendix~\ref{app:data}), which is the input a judge sees in practice.
Deployed agents stack cut context the same way: the OpenAI Agents SDK provides a trimmer that cuts tool outputs above 500 characters outside the last two turns, and Anthropic's guidance for long-running agents clears old tool results and compacts earlier turns \citep{openai2026trimmer,rajasekaran2025context}.
FCRB requests have no earlier calls, so nothing is cut there.

\smallskip  \noindent \textbf{Statistics.}
All confidence intervals (CIs) are 95\% bootstrap intervals over 1,000 resamples of queries.
A difference is called significant when its CI excludes zero (Appendix~\ref{app:metrics}).

\subsection{Measuring the Gap}
\label{sec:gap}

\begin{table*}[!t]
  \centering
  % \footnotesize
  \scriptsize
  \begin{tabular}{lcccccc}
    \toprule
    & \multicolumn{2}{c}{FCRB} & \multicolumn{2}{c}{CFB} & \multicolumn{2}{c}{FinTrace} \\
    \cmidrule(lr){2-3}\cmidrule(lr){4-5}\cmidrule(lr){6-7}
    Judge & Out of view & Default & Out of view & History-only & Out of view & No-source \\
    \midrule
    CLM & \makecell{8.15\\[-1pt]{\scriptsize[3.92, 12.07]}} & \makecell{31.93\\[-1pt]{\scriptsize[24.36, 41.20]}} & \makecell{30.76\\[-1pt]{\scriptsize[27.95, 34.02]}} & \makecell{66.63\\[-1pt]{\scriptsize[59.42, 73.00]}} & \makecell{25.95\\[-1pt]{\scriptsize[23.46, 28.22]}} & \makecell{28.28\\[-1pt]{\scriptsize[25.34, 30.81]}} \\
    Jev & \makecell{8.11\\[-1pt]{\scriptsize[4.44, 11.74]}} & \makecell{20.26\\[-1pt]{\scriptsize[12.68, 27.32]}} & \makecell{22.67\\[-1pt]{\scriptsize[19.08, 27.71]}} & \makecell{59.39\\[-1pt]{\scriptsize[54.22, 68.00]}} & \makecell{27.78\\[-1pt]{\scriptsize[25.20, 29.87]}} & \makecell{27.47\\[-1pt]{\scriptsize[24.54, 30.25]}} \\
    \midrule
    Qwen3-4B & \makecell{4.37\\[-1pt]{\scriptsize[0.37, 8.43]}} & \makecell{21.26\\[-1pt]{\scriptsize[13.91, 29.63]}} & \makecell{34.71\\[-1pt]{\scriptsize[32.87, 36.78]}} & \makecell{76.81\\[-1pt]{\scriptsize[71.30, 81.93]}} & \makecell{23.30\\[-1pt]{\scriptsize[20.91, 25.94]}} & \makecell{23.29\\[-1pt]{\scriptsize[20.73, 26.35]}} \\
    Qwen3-8B & \makecell{0.63\\[-1pt]{\scriptsize[$-$3.78, 4.35]}} & \makecell{14.40\\[-1pt]{\scriptsize[4.69, 22.14]}} & \makecell{33.30\\[-1pt]{\scriptsize[31.13, 35.65]}} & \makecell{74.87\\[-1pt]{\scriptsize[68.09, 80.47]}} & \makecell{24.04\\[-1pt]{\scriptsize[22.13, 26.22]}} & \makecell{25.72\\[-1pt]{\scriptsize[23.59, 28.31]}} \\
    Qwen3.8-27B & \makecell{12.29\\[-1pt]{\scriptsize[8.19, 15.81]}} & \makecell{36.03\\[-1pt]{\scriptsize[27.56, 44.48]}} & \makecell{25.99\\[-1pt]{\scriptsize[21.10, 30.72]}} & \makecell{57.08\\[-1pt]{\scriptsize[50.28, 63.87]}} & \makecell{29.76\\[-1pt]{\scriptsize[27.44, 32.10]}} & \makecell{31.68\\[-1pt]{\scriptsize[28.98, 34.22]}} \\
    \midrule
    Llama-3.1-8B & \makecell{8.31\\[-1pt]{\scriptsize[4.54, 12.33]}} & \makecell{49.00\\[-1pt]{\scriptsize[40.85, 56.87]}} & \makecell{23.95\\[-1pt]{\scriptsize[19.54, 28.02]}} & \makecell{55.97\\[-1pt]{\scriptsize[50.37, 62.85]}} & \makecell{23.31\\[-1pt]{\scriptsize[20.53, 25.64]}} & \makecell{23.54\\[-1pt]{\scriptsize[20.25, 26.28]}} \\
    Mistral-7B & \makecell{12.25\\[-1pt]{\scriptsize[8.33, 16.03]}} & \makecell{50.51\\[-1pt]{\scriptsize[42.23, 58.31]}} & \makecell{36.13\\[-1pt]{\scriptsize[33.67, 38.43]}} & \makecell{80.54\\[-1pt]{\scriptsize[73.82, 86.99]}} & \makecell{30.68\\[-1pt]{\scriptsize[28.61, 32.85]}} & \makecell{32.47\\[-1pt]{\scriptsize[30.29, 34.82]}} \\
    \bottomrule
  \end{tabular}
  \caption{Derivation gap $\mathrm{Gap}_\tau$ (in points) with Plain input, with its 95\% CI below. Out of view pools the four paths other than \textsc{Visible}. The second column of each dataset is its main path.}
  \label{tab:gap}
\end{table*}

\smallskip  \noindent \textbf{Metric.}
We compare how often a judge accepts correct values on different derivation paths.
A judge accepts a value when its score is above a threshold.
One fixed threshold for all judges is not fair, because judges accept values at very different score levels.
On CFB, for example, Qwen3-8B accepts 80\% of correct arguments at a score of 0.002, and CLM at a score of 0.631.
For each judge, dataset, and input, we therefore set a threshold $\tau$ at which the judge accepts 80\% of correct arguments.
Let $A_\tau(p)$ be the share of correct arguments on path $p$ that the judge accepts at $\tau$.
The \emph{derivation gap} of path $p$ is
\begin{equation}
  \mathrm{Gap}_\tau(p) = A_\tau(\textsc{Visible}) - A_\tau(p).
  \label{eq:gap}
\end{equation}
A positive gap means that correct values on $p$ are accepted less often than correct \textsc{Visible} values.
We say that values on the \textsc{Visible} path are \emph{in view} and that values on the other four paths are \emph{out of view}.
We also report the gap of the out-of-view values pooled together.
% The gap compares acceptance rates and not AUROC within a path.
% A deployed judge applies one threshold to every argument, so a lower acceptance rate on a path means more false rejections on that path.

The gap compares acceptance rates and not AUROC within a path. This is because AUROC within a path can stay the same when a judge scores all values on that path lower, right and wrong alike. A deployed judge applies one threshold to every argument, so such a shift means more false rejections on that path.
Appendix~\ref{app:metrics} gives the rule for ties at $\tau$.
It also repeats the analysis with $\tau$ set to accept 70\% or 90\%, where every CFB and FinTrace gap tested stays significant.

\smallskip  \noindent \textbf{Results.}
Table~\ref{tab:gap} gives the gap for all seven judges.
For each benchmark, it reports the pooled out-of-view gap and the gap on the main out-of-view path of that benchmark: \textsc{Default} on FCRB, \textsc{History-only} on CFB, and \textsc{No-source} on FinTrace.
On CFB and FinTrace, the out-of-view gap is significant for every judge.
On FCRB it is small, and defaults carry the largest gap.
Scale does not remove the gap, since Qwen3.8-27B, the largest judge we run, has a significant out-of-view gap on all three benchmarks, between 12.29 and 29.76 points.

\smallskip  \noindent \textbf{Base rates.}
A gap could also arise from the base rate of a path, the proportion of its values that are correct: if the base rate of a path is lower, a judge would be right to trust its values less.
On FCRB, 77\% of default values are correct, against 75\% of values in view.
On CFB, 83\% of history-only values are correct, against 82\% in view.
So base rates do not explain the gap in either case.
On FinTrace, only 45\% of no-source values are correct, against 97.5\% in view, so this low base rate may account for part of the gap.

\smallskip  \noindent \textbf{Truncation is not the whole story.}
On CFB, truncation is an obvious cause of the history-only gap, because it removes the text that these values match.
But it cannot be the only cause.
On FCRB nothing is cut, the default is written in the schema the judge reads, and correct defaults still carry the largest gap of any path.
A judge might reasonably hesitate over a default that the user never mentioned.
Yet base rates do not explain this gap.
This suggests a second explanation besides missing information: the input has no explicit reference, that is, no text that states what the value should be.
Section~\ref{sec:follow} examines both explanations with controlled tests.

\subsection{Judges Follow the Reference}
\label{sec:follow}
\begin{figure}[t]
  \centering
  \includegraphics[width=\columnwidth]{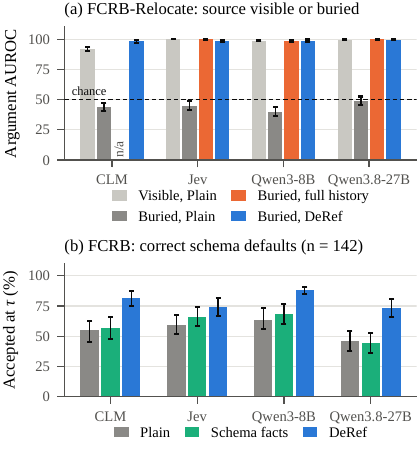}
  \caption{(a) Argument AUROC on FCRB-Relocate, with the source of the correct value in view (visible) or past the cut (buried). CLM is run at its default input length of 2,048 tokens and is not given the full history. (b) Acceptance at $\tau$ of the 142 correct FCRB defaults whose \method{} reference says the default applies. Schema facts restates the default and lists related words in the request. Error bars show 95\% CIs.}
  \label{fig:follow}
\end{figure}
Two explanations remain for a rejected correct value.
The judge may lack the information.
Or it may have the information but no explicit reference.
The tests below examine these two explanations.
They use the four judges named in Section~\ref{sec:gap-setup}.

\smallskip  \noindent \textbf{Test 1: judges prefer visible distractors when sources are hidden.}
If a judge merely lacked information, hiding the source should push its score toward uncertainty.
Instead it follows the wrong value it can see.
We build FCRB-Relocate from 257 FCRB states.
We move the value of one argument from the user request into the return of an added earlier call.
The request now points to the right entry by one of its attributes.
The return lists eight entries, the correct one and seven distractors whose values come from other FCRB samples.
Each state has two versions that differ only in the order of the entries.
The correct entry comes second, within the first 1,000 characters that the judge sees, or last, past that cut.
The first distractor stays in view in both versions.
This mirrors how history-only values arise in CFB (Appendix~\ref{app:data}).
Burying the source drops argument AUROC from between 91.89 and 99.88 to between 39.80 and 49.03 for all four judges (Figure~\ref{fig:follow}a).
Against the distractor in view alone, AUROC is between 26.71 and 39.88, so all four judges rank the wrong value they can see above the correct value that was cut.
Given the full history, Jev, Qwen3-8B, and Qwen3.8-27B recover, to between 98.39 and 99.86.

\smallskip  \noindent \textbf{Test 2: restoring the information closes the gap only where a source was cut.}
If truncation were the whole cause, the uncut history should close the gap on every path.
We give Jev, Qwen3-8B, and Qwen3.8-27B the full history, with returns uncut and no window, up to each judge's input limit (Appendix~\ref{app:data}).
On CFB, this narrows the gap on history-only values by between 37.50 and 62.84 points, and for Qwen3.8-27B the remaining gap is no longer significant (Table~\ref{tab:grid-gap} in Appendix~\ref{app:grid}).
Argument AUROC on these values rises by between 11.45 and 25.95 points.
But on no-source values it raises AUROC for none of the three judges, and it lowers it for Jev by 2.83 points and for Qwen3-8B by 13.83 points (Table~\ref{tab:grid-cfb} in Appendix~\ref{app:grid}).
Putting back only the one cut step that holds the source has the same effect on FinTrace: it raises the scores of correct history-only values for all four judges, against a placebo step of the same kind (Appendix~\ref{app:data}).
A reference also helps on top of the uncut history (Section~\ref{sec:paths}).

\smallskip  \noindent \textbf{Test 3: the information can be in view and still not count.}
In FCRB the default is written in the parameter description that the judge sees, and nothing is cut, yet judges accept correct defaults less often than correct values in view (Table~\ref{tab:gap}).
On the 142 of the 185 correct defaults whose \method{} reference (Section~\ref{sec:method}) says that the default applies, a line that restates the schema default and lists which related words appear in the request changes acceptance by at most 6.18 points.
The reference itself, which writes out this derivation, raises it to between 73.02 and 88.06 (Figure~\ref{fig:follow}b).
This subset is chosen by the reference.
Table~\ref{tab:grid-gap} in Appendix~\ref{app:grid} reports all 185, with the same picture.

\smallskip  \noindent \textbf{A matching reference lifts wrong values too.}
The three tests say that judges need a reference.
The last test shows that judges follow a reference even when it supports a wrong value.
On CFB, we add snippets from the full history around each place where the candidate value occurs, built by the snippet rule of Appendix~\ref{app:more}.
Such snippets contain the value by construction, so they agree with any candidate.
On the 59 wrong history-only arguments, they raise the mean score for all four judges, by between 14.28 and 39.39 points.
At the 0.5 decision threshold, CLM, Jev, and Qwen3-8B then accept between 28 and 49 of the 59, while Qwen3.8-27B accepts 2.
On FinTrace-HC, which pairs for the same argument a wrong value that appeared earlier in the history (H) with one that did not (C), CLM, Jev, and Qwen3-8B score H higher than C (Appendices~\ref{app:data} and~\ref{app:more}).

These tests show one pattern.
A judge scores a value higher when a reference in its input matches it, whether the value is right or wrong.
It does not reliably rebuild the derivation on its own, even from the full history.

\section{\method{}: Deriving the Reference}
\label{sec:method}
\method{} gives the judge a reference that is ready to use.
It has two steps (Figure~\ref{fig:overview}b).
A deriver model reads the full history $H$ and the schema, with returns uncut and no window (Appendix~\ref{app:impl}).
The deriver sees the function called at the step under judgment and its schema, but no candidate value.
For each argument it writes the expected value and its source.
If the history does not fix the value, it says so, and that statement becomes the reference.
Our deriver reads everything, so it can name values whose source was cut from the judge input (Section~\ref{sec:paths}).
Because the deriver never sees the candidate, the reference cannot echo the value under test.
The deriver writes one reference per argument of a state, and all candidates at that state share it.
We use Qwen3.8-27B \citep{qwenteam2026qwen3827b} as the deriver for every judge, with thinking on (Appendix~\ref{app:more} gives the prompt and Appendix~\ref{app:impl} the settings).
We insert the reference into the judge input, after the history and before the function description.
The judge itself is unchanged and scores each argument as before.
In all main experiments, every judge keeps its usual truncated input and reads the reference in addition (Appendix~\ref{app:data}).

% ----------------------------------------------------------------------------
% Section 3.4 (Judges Follow the Reference): the old Section 3.3
% How Victor's comments were addressed
% (35) "Clarify that this is in the appendix": the reference now reads
%     "Table 17 in Appendix H".
% (36) "Same here, check this for all referenced tables": done for the three
%     appendix tables cited in this section (Table 17 twice, Table 18 once).
%     Appendix tables cited in Section 5 are still to be checked.
% (37) "What is 'it' here?": the sentence now reads "judges follow a
%     reference even when it supports a wrong value". 
% ----------------------------------------------------------------------------

\section{Experiments}
\label{sec:exp}
We ask whether \method{} improves every judge and how it compares with prior ways to give the judge evidence \textbf{(RQ1)}, where the gains come from \textbf{(RQ2)}, and whether the reference drives the judgment \textbf{(RQ3)}.

\subsection{Setup}
\label{sec:setup}
\paragraph{Data and judges.}
We use the three benchmarks of Section~\ref{sec:gap-setup} and the two controlled sets (Section~\ref{sec:follow} and Appendix~\ref{app:data}).
Our main experiments run \method{} with four of the seven judges: CLM, Jev, Qwen3-8B, and Qwen3.8-27B.
Qwen3-8B is the frozen backbone of CLM, so the two share one base model.
Qwen3.8-27B is the largest open LLM we run.

\smallskip  \noindent \textbf{Inputs.}
Plain is the truncated input of Section~\ref{sec:gap-setup}.
We compare \method{} with three prior methods (Appendix~\ref{app:more}).
ProvenanceGuard evidence gives the judge snippets of the history that a model quotes as the source of the proposed value \citep{she2026safeguarding}.
RevisEval gives the judge a revised version of the proposed call \citep{zhang2025reviseval}.
Both use Qwen3.8-27B with the same settings as our deriver, and both see the proposed value.
Self-generated reference has the judge model write the expected value from its own truncated input before it judges \citep{zeng2024llmbar}.
It needs a judge that writes text, so it runs with Qwen3-8B and Qwen3.8-27B.
Full history has no window and no cut, up to about 32K tokens (FCRB has no cut).
We run it with Jev and the two LLM judges. We report argument AUROC and $\mathrm{Gap}_\tau$ with $\tau$ set again for each input. Appendix~\ref{app:call} gives call-level AUROC.

\begin{table}[t]
  \centering
  \scriptsize
  \setlength{\tabcolsep}{3pt}
  \renewcommand{\arraystretch}{0.92}
  \begin{tabular}{lccc}
    \toprule
    Judge and evidence & FCRB & CFB & FinTrace \\
    \midrule
    \multicolumn{4}{l}{\textit{CLM}} \\
    Plain & 72.51 & 63.41 & 83.36 \\
    \makecell[l]{ProvenanceGuard evidence\\[-1pt]{\scriptsize\citep{she2026safeguarding}}} & 75.17 & 69.62 & 84.75 \\
    \makecell[l]{RevisEval\\[-1pt]{\scriptsize\citep{zhang2025reviseval}}} & 83.26 & 81.78 & 86.07 \\
    \method{} (ours) & \textbf{89.59} & \textbf{85.55} & \textbf{89.10} \\
    $\Delta$ vs.\ RevisEval & \makecell{+6.33\\[-1pt]{\scriptsize[4.65, 7.98]}} & \makecell{+3.77\\[-1pt]{\scriptsize[1.72, 5.90]}} & \makecell{+3.03\\[-1pt]{\scriptsize[1.42, 4.79]}} \\
    \midrule
    \multicolumn{4}{l}{\textit{Jev}} \\
    Plain & 90.83 & 75.01 & 89.11 \\
    Full history & -- & 79.82 & 90.08 \\
    ProvenanceGuard evidence & 91.00 & 80.74 & 89.57 \\
    RevisEval & 92.23 & 87.15 & 91.77 \\
    \method{} (ours) & \textbf{93.30} & \textbf{88.73} & \textbf{93.10} \\
    $\Delta$ vs.\ RevisEval & \makecell{+1.07\\[-1pt]{\scriptsize[$-$0.05, 2.13]}} & \makecell{+1.57\\[-1pt]{\scriptsize[0.31, 2.98]}} & \makecell{+1.33\\[-1pt]{\scriptsize[0.77, 1.88]}} \\
    \midrule
    \multicolumn{4}{l}{\textit{Qwen3-8B}} \\
    Plain & 78.79 & 66.79 & 85.15 \\
    Full history & -- & 66.92 & 84.74 \\
    \makecell[l]{Self-generated reference\\[-1pt]{\scriptsize\citep{zeng2024llmbar}}} & 81.80 & 68.28 & 87.95 \\
    ProvenanceGuard evidence & 77.71 & 64.80 & 85.21 \\
    RevisEval & 87.06 & 81.75 & 90.52 \\
    \method{} (ours) & \textbf{89.48} & \textbf{83.13} & \textbf{92.28} \\
    $\Delta$ vs.\ RevisEval & \makecell{+2.42\\[-1pt]{\scriptsize[0.89, 4.01]}} & \makecell{+1.38\\[-1pt]{\scriptsize[$-$0.64, 3.26]}} & \makecell{+1.75\\[-1pt]{\scriptsize[0.97, 2.53]}} \\
    \midrule
    \multicolumn{4}{l}{\textit{Qwen3.8-27B}} \\
    Plain & 91.95 & 78.92 & 89.31 \\
    Full history & -- & 85.45 & 89.77 \\
    Self-generated reference & 92.52 & 81.83 & 90.05 \\
    ProvenanceGuard evidence & 91.94 & 88.08 & 90.09 \\
    RevisEval & \textbf{93.70} & 93.54 & \textbf{92.53} \\
    \method{} (ours) & 93.39 & \textbf{95.09} & 92.25 \\
    $\Delta$ vs.\ RevisEval & \makecell{$-$0.32\\[-1pt]{\scriptsize[$-$1.48, 0.80]}} & \makecell{+1.55\\[-1pt]{\scriptsize[0.25, 2.83]}} & \makecell{$-$0.27\\[-1pt]{\scriptsize[$-$0.93, 0.38]}} \\
    \bottomrule
  \end{tabular}
  \caption{Argument AUROC (in points) of four judges with each kind of evidence (Section~\ref{sec:setup}). Bold marks the best input for each judge and benchmark. $\Delta$ is \method{} minus RevisEval, with its 95\% CI below. FCRB has no cut, so Full history equals Plain there. Table~\ref{tab:grid-main} adds \method{}-v and the gain over Plain.}
  \label{tab:main}
\end{table}

\subsection{Main Results}
\label{sec:main}

\smallskip  \noindent \textbf{\method{} improves every judge.}
Table~\ref{tab:main} answers RQ1.
It covers four judges and three benchmarks, that is, 12 pairs of judge and benchmark.
\method{} raises argument AUROC over Plain in all 12 pairs, by between 1.44 and 22.14 points, and every gain is significant (Table~\ref{tab:grid-main} in Appendix~\ref{app:grid}).
The Llama and Mistral judges gain between 8.78 and 20.84 points, also significantly (Table~\ref{tab:other-judges} in Appendix~\ref{app:more}).
The gains carry over to call-level AUROC, where all 12 are again significant (Appendix~\ref{app:call}).

\smallskip  \noindent \textbf{Comparison with prior methods.}
RevisEval is the strongest prior method in all 12 pairs.
\method{} is above it in 10 pairs and significantly above it in 8.
Averaged over the four judges, \method{} is significantly above RevisEval on every benchmark (Appendix~\ref{app:grid}).
In the other two pairs, Qwen3.8-27B on FCRB and FinTrace, RevisEval is higher by 0.32 and 0.27 points, and neither is significant.
\method{} is significantly above ProvenanceGuard evidence in all 12 pairs.
Self-generated reference gains at most 3.01 points over Plain and stays below \method{} in all six pairs where it runs.

\smallskip  \noindent \textbf{More context and more reasoning.}
Giving the judge the full history instead of a reference is significantly below \method{} (Table~\ref{tab:main}).
Averaged over the three judges that are given it, the difference is 11.58 points on CFB and 4.35 on FinTrace (Appendix~\ref{app:grid}).
More reasoning by the judge does not make up for the reference either.
Qwen3.8-27B judging directly with thinking on stays below the same model reading the reference with thinking off on all three benchmarks, by between 2.82 and 10.92 points (Appendix~\ref{app:more}).

\smallskip  \noindent \textbf{Controlled sets.}
On the buried version of FCRB-Relocate, where all four judges fell to chance or below (Section~\ref{sec:follow}), \method{} brings them back to between 98.40 and 99.59 (Figure~\ref{fig:follow}a).
This includes CLM, which is not given the full history.
On the visible version and on FinTrace-HC, where the judges were already accurate, \method{} makes no judge significantly worse (Table~\ref{tab:controlled} in Appendix~\ref{app:data}).

% ----------------------------------------------------------------------------
% Section 5.2 (Main Results)
% How Victor's comments were addressed
% (38) "Give structure to the main results by type of result": the single
%     long paragraph is split into four named paragraphs (improvement over
%     Plain, comparison with prior methods, more context and more reasoning,
%     controlled sets).
% (39) "All 12 what?": the first paragraph now says that Table 2 covers four
%     judges and three benchmarks, that is, 12 pairs of judge and benchmark.
% (40) "Cells? What are cells?": "cells" is replaced by "pairs" throughout.
% (41) "Very hard to parse, break it in two": the full-history sentence is
%     split into two.
% (42) "Again this sentence is very very long": the DeRef-v comparison is
%     moved to the ablation (Section 5.3), where it belongs, and reduced to
%     two sentences. The details are in Appendix G.
% (36) "Check appendix references for all tables": every appendix table cited
%     in this section now reads "Table X in Appendix Y".

\subsection{Ablation: Where the Gain Comes From}
\label{sec:paths}

\begin{table}[!t]
  \centering
  \scriptsize
  \setlength{\tabcolsep}{4pt}
  \begin{tabular}{lcc}
    \toprule
    Variant & AUROC & $\Delta$ \\
    \midrule
    \method{} & 95.09 & -- \\
    w/o full history & 85.46
      & \makecell{+9.63\\[-1pt]{\scriptsize[7.53, 11.86]}} \\
    Qwen3-8B as deriver & 88.56
      & \makecell{+6.53\\[-1pt]{\scriptsize[4.70, 8.39]}} \\
    Deriver thinking off & 88.18
      & \makecell{+6.91\\[-1pt]{\scriptsize[5.42, 8.47]}} \\
    Parsed expected value only & 90.90
      & \makecell{+4.19\\[-1pt]{\scriptsize[3.15, 5.43]}} \\
    \bottomrule
  \end{tabular}
  \caption{Ablation of \method{} on CFB with Qwen3.8-27B (argument AUROC in points).
  Each variant changes one part of \method{}.
  The judge reads the truncated input with thinking off throughout.
  $\Delta$ is \method{} minus the variant, with its 95\% paired bootstrap CI.
}
  \label{tab:ablation}
\end{table}

\begin{table}[!t]
  \centering
  \scriptsize
  \setlength{\tabcolsep}{4pt}
  \begin{tabular}{lccc}
    \toprule
    & \multicolumn{2}{c}{Reference} & \\
    \cmidrule(lr){2-3}
    Judge thinking & Without & With & $\Delta_{\mathrm{ref}}$ \\
    \midrule
    Off & 85.45 & 96.01
      & \makecell{+10.56\\[-1pt]{\scriptsize[8.82, 12.26]}} \\
    On & 90.27 & 94.74
      & \makecell{+4.47\\[-1pt]{\scriptsize[2.74, 6.33]}} \\
    \midrule
    $\Delta_{\mathrm{think}}$
      & \makecell{+4.82\\[-1pt]{\scriptsize[3.18, 6.46]}}
      & \makecell{$-$1.26\\[-1pt]{\scriptsize[$-$1.86, $-$0.79]}}
      & -- \\
    \bottomrule
  \end{tabular}
  \caption{Reference availability and judge thinking on CFB with Qwen3.8-27B (argument AUROC in points).
  All four settings give the judge the full history.
  The two settings with a reference reuse the same references, written without the candidate and with deriver thinking on.
  $\Delta_{\mathrm{ref}}$ is With minus Without within each row, and $\Delta_{\mathrm{think}}$ is On minus Off within each column, with 95\% paired bootstrap CIs.}
  \label{tab:reference-thinking}
\end{table}

Table~\ref{tab:ablation} removes one part of \method{} at a time.
We show CFB with Qwen3.8-27B because CFB has the largest gap on a single path and includes sources that were cut from the judge input.
Every drop is significant, and the largest comes from writing the reference from the judge's cut input instead of the full history.

\smallskip  \noindent \textbf{Why keep the judge.}
Keeping only the parsed expected value costs little where the reference states a value (1.39 points) and much where the parser finds none (22.37 points), so the text beyond the stated value helps mainly where it still carries the value (Appendix~\ref{app:parser}).
This is also why we keep the judge instead of matching the candidate against the parsed value: where the parser finds no value, a string match has nothing to compare, and the judge still uses the rest of the reference.

\smallskip  \noindent \textbf{Reference or more thinking.}
Table~\ref{tab:reference-thinking} crosses the reference with judge thinking, with the full history in all four settings.
The best setting is the reference with thinking off, at 96.01.
So the gain comes from the reference, not from more reasoning by the judge.

\smallskip  \noindent \textbf{Why hide the candidate.}
A deriver that sees the candidate, \method{}-v, is above \method{} on CFB and below it on FCRB and FinTrace, and it names the wrong value under test as the expected value up to six times as often (Appendix~\ref{app:more}).
We therefore keep the deriver that does not see the value.

\subsection{Changing the Reference Changes the Judgment}
\label{sec:swap}
\begin{figure}[t]
  \centering
  \includegraphics[width=0.9\columnwidth]{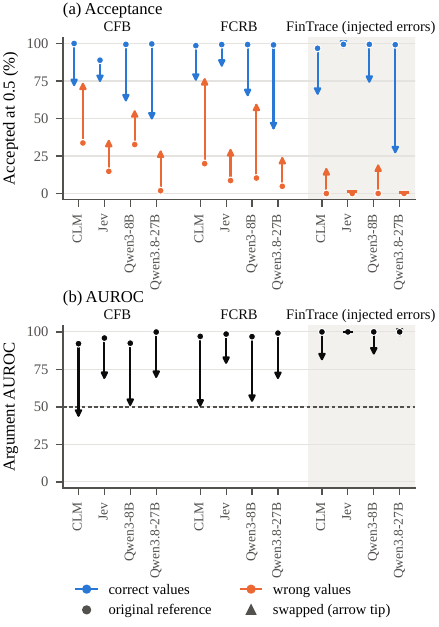}
  \caption{Changing one value in the reference. Dots show the original reference, which names the correct value. Arrow heads show it with a wrong value from the same state.}
  \label{fig:swap}
\end{figure}
We now test whether judges follow the \method{} reference itself.
We take up to 500 arguments per benchmark whose reference names the correct value (332 on CFB, 417 on FCRB, and 500 on FinTrace).
In the reference text, we replace that value with a wrong value for the same argument from the same state and keep all other text.
On CFB and FCRB, where the wrong values come from real models, all four judges follow the change (Figure~\ref{fig:swap}).
At the 0.5 threshold, acceptance of correct values drops by between 15.36 and 57.07 points, and acceptance of wrong values rises by between 20.38 and 57.79 points.
Argument AUROC falls from between 92.16 and 99.91 to between 42.63 and 77.92.
All 24 changes are significant.
The swap also overturns judgments that agreed with the labels under Plain.
For correct values that a judge accepted with Plain input, acceptance drops by between 10.85 and 53.62 points.
For values labeled wrong that it rejected, acceptance rises by between 11.00 and 50.74 points.
FinTrace behaves differently, because we made its wrong values by rule, and under Plain the judges accept at most 11.44\% of them at 0.5.
After the swap, their acceptance rises by at most 20.20 points, and AUROC stays between 80.26 and 99.77.
Still, CLM, Qwen3-8B, and Qwen3.8-27B lower their acceptance of correct values by between 26.40 and 73.20 points, and Jev by 3.00 points.
Since the reference is the only text we change, the change in judgment comes from the reference (examples in Appendix~\ref{app:cases}).

\smallskip  \noindent \textbf{Summary.} RQ1: \method{} improves argument AUROC for every tested judge on every benchmark and significantly outperforms RevisEval when averaged over judges. RQ2: deriving a reference from the full history provides gains that more context or judge thinking alone does not achieve. RQ3: swapping the reference to name a wrong value can reverse judgments on CFB and FCRB.

\section{Conclusion}
Argument judges tend to score the candidate against explicit reference values in their input.
When a correct value matches only a return that was cut, or is a default with nothing saying that it applies, the input holds no such reference, and judges of every type and size we test accept it less often than a value shown in the input.
On the two multi-step benchmarks, the same holds for values that match no span of the history.
\method{} derives the reference from the full history before the candidate is seen.
Providing this reference improves overall argument AUROC for all six judges we test on three benchmarks, with no change to the judges, and substantially reduces the gap on cut returns and on defaults.

\section*{Limitations}
Each argument has one gold value in our data, so a reference that names it can lead judges to reject another value that is also correct (Appendix~\ref{app:data}). On CFB and FinTrace, the wrong labels come from our own construction, and an audit of 100 items per dataset by two annotators could not confirm 3\% and 4\% of them. Appendix~\ref{app:data} runs a worst-case analysis of this label uncertainty, and every gain of \method{} over Plain remains significant. FCRB labels follow the official BFCL checker and are not independently audited.

On FinTrace, \method{} does not shrink the gap on no-source values for three of four judges (Table~\ref{tab:grid-gap}). The low base rate of this path may account for part of that gap (Section~\ref{sec:gap}).

Main results use one deriver model, Qwen3.8-27B, held fixed across judges and input conditions. In the ablation (Table~\ref{tab:ablation}), a weaker deriver, Qwen3-8B, still improves Qwen3.8-27B's argument AUROC over Plain on CFB. Broader evaluation of derivers from other model families is left to future work.

Our judges include dedicated scoring models (CLM and Jev) and open LLMs of up to 27B parameters, which show the pattern across judge types and sizes. Evaluation on general-purpose frontier LLMs as judges is left to future work.

% Each argument has one gold value in our data, so a reference that names it can lead judges to reject another value that is also correct (Appendix~\ref{app:data}).
% On CFB and FinTrace, the wrong labels come from our own construction, and an audit of 100 items per dataset by two annotators could not confirm 3\% and 4\% of them.
% Appendix~\ref{app:data} bounds the effect of this noise, and every gain of \method{} over Plain stays significant under it.
% FCRB labels follow the BFCL checker and are not audited, yet a deriver that never sees the candidate still names the labeled wrong value for 8.19\% of wrong FCRB arguments, so some of these labels may also be too strict.

% On FinTrace, \method{} does not shrink the gap on no-source values for three of four judges (Table~\ref{tab:grid-gap} in Appendix~\ref{app:grid}).

% All main results use one deriver model, Qwen3.8-27B, which is also one of the judges.
% We hold the deriver fixed so that differences across judges and inputs reflect the judge and the input alone.
% A weaker deriver, Qwen3-8B, still lifts Qwen3.8-27B by 9.64 points over Plain on CFB (Table~\ref{tab:ablation}), so the effect does not depend on this one deriver, but its references may still favor judges of its own family.
% Derivers outside the Qwen family are left to future work.

% Our largest open judge is Qwen3.8-27B, and Jev is a closed source judge of unknown size, so we test no general-purpose frontier LLM as a judge.

\section*{Ethical Considerations}
We use three public benchmarks as released, together with public trajectories of agent runs on ComplexFuncBench.
Their requests ask for function calls in tasks such as travel booking and financial queries, and their tool returns hold data such as hotel listings and company financials.
We do not collect new data from people, and we do not try to identify anyone.
Names in the requests come from the benchmarks, and we add no personal information.
Two authors checked samples of labels and paths by hand, and we used no crowd workers.
Jev, which we reach through an outside API, receives only benchmark text and model outputs.

Argument judges can gate tool calls that have real effects, such as a booking or a trade.
Our results show that judges follow the reference in their input, and the swap test shows that a wrong reference overturns many judgments.
So a deriver that errs, or a tool return written to mislead it, could make a judge accept a wrong argument.
We therefore see \method{} as support for review, not a replacement for it.
Calls with high stakes should still go to a person or to a check that does not rely on the judge.
We recommend keeping each derived reference with its scores, so that a reviewer can see why a value was accepted.

The benchmarks are task-oriented, and their texts are function requests, schemas, and tool returns.
We did not see offensive content in the items we read for our label and path checks.
We did not screen every item, and we use the texts as released.

\section*{Artifact Use and Licensing}
FC-RewardBench and ComplexFuncBench are released under Apache-2.0, and FinTrace under MIT.
CLM, Qwen3, Qwen3.8-27B, and Mistral-7B-Instruct-v0.3 are released under Apache-2.0, and vLLM under Apache-2.0.
We use Llama-3.1-8B-Instruct under the Llama 3.1 Community License.
We access Jev through OpenRouter under its terms of service.
After review, we will release our code, the path labels, and the two controlled sets.

\section*{Use of AI Assistants}
The authors used AI assistants, including ChatGPT, to support language polishing, brainstorming, and LaTeX editing.
All technical claims, experimental results, analysis, and final writing decisions were reviewed and verified by the authors.

\bibliography{custom}

\clearpage
\appendix
\renewcommand{\topfraction}{0.9}\renewcommand{\dbltopfraction}{0.9}\renewcommand{\bottomfraction}{0.8}
\renewcommand{\textfraction}{0.05}\renewcommand{\floatpagefraction}{0.8}\renewcommand{\dblfloatpagefraction}{0.8}
\setcounter{topnumber}{4}\setcounter{dbltopnumber}{4}\setcounter{bottomnumber}{2}\setcounter{totalnumber}{6}

\begin{figure*}[!htbp]
\centering
\begin{minipage}{0.97\textwidth}
\footnotesize
\begin{verbatim}
Question: Generate a digital image with captured 'a man wearing a red dress'.
Previous calls (most recent 5):
none
Function: generate_human_image -- Generates a digital image of a human subject
  based on the provided text prompt. This function is tailored to create images
  representing different human demographics such as girls, boys, women, men,
  and children.
Parameter: image_quality -- The desired quality level of the generated image.;
  type: string; enum: ["low", "medium", "high"]; default: "high"
Proposed value for image_quality: "high"
\end{verbatim}
\end{minipage}
\caption{A Plain judge input from FCRB, a correct value on the \textsc{Default} path. Long lines are wrapped here for space. The judge question is ``Is this the correct value for this parameter, given the question?'' \method{} inserts its block after the state, before the \texttt{Function} line.}
\label{fig:input}
\end{figure*}
\section{Data and Judge Inputs}
\label{app:data}
\paragraph{Choice of benchmarks.}
We need three things: multi-step trajectories with real tool returns, so that sources can be cut, a gold value for each argument, and wrong calls made by real models.
TRBench\_BFCL uses the same BFCL tasks as FCRB in its single-turn part and compares whole turns of calls in its multi-turn part \citep{li2026toolrm}.
The data of ProvenanceGuard label calls for safety or judge tasks by their final outcome, so they give no gold value per argument \citep{she2026safeguarding}.

\paragraph{FCRB.}
From the 1,500 single-turn requests of FC-RewardBench we drop the 15 with unparseable wrong outputs.
We keep the 650 whose wrong call fails the value check of BFCL (error type \texttt{value\_error}).
We then remove 20 whose only recorded message is a system prompt.
We check each argument of the wrong call on its own with the BFCL checker.
It is labeled wrong when the checker rejects its value and carries no label otherwise.
This leaves 630 requests, each with one correct and one wrong call, and 1,719 correct and 769 wrong argument values.
FCRB has no earlier calls, so the judge sees the request only.

\paragraph{CFB.}
We use the released trajectories of 30 models on the 1,000 CFB tasks (revision \texttt{fcc906da}).
For each trajectory we find the first step where the model call differs from the gold call.
We keep the divergences in which the model calls the gold function with different argument values, since only these give an argument to label; a call to another function or a missing call is left out.
The gold history up to that step is the state.
The gold call is the correct call and the model call is the wrong call.
Models that make the same wrong call at a state are counted once, so a state can carry several wrong calls.
Parallel calls are matched to gold calls by position.
We drop 30 wrong argument values whose parallel calls to the same function could not be matched in order.
This leaves 347 correct and 644 wrong calls, with 1,639 correct and 999 wrong argument values.

\paragraph{FinTrace.}
FinTrace gives 800 test queries with golden trajectories \citep{cao2026fintrace}. It labels whole trajectories but no wrong calls at the argument level.
We make wrong calls from golden calls by rule, with a fixed seed (20260930).
Each wrong call keeps the endpoint and changes one argument.
A ticker becomes a ticker used in another query, a date or year moves by one to three years (a \texttt{from} and \texttt{to} pair moves together), or the reporting period switches between annual and quarter.
We keep calls whose function descriptions come from the documentation of the data provider, and we drop 10 calls whose request has an unfilled template.
This leaves 2,334 correct and 2,334 wrong calls, with 7,428 correct and 2,796 wrong argument values.

\paragraph{Label checks and sensitivity analysis.}
Two annotators independently checked 100 random arguments labeled wrong from each of CFB and FinTrace, without seeing any judge score or derived reference.
They agreed on 94\% of the CFB items and 88\% of the FinTrace items, and settled the rest by discussion.
Of 100 CFB arguments, 97 were confirmed wrong, 2 could also be correct, and 1 was unclear.
Of 100 FinTrace arguments, 96 were confirmed wrong, 2 could also be correct, and 2 were unclear.
We checked CFB and FinTrace because their argument labels come from our construction procedures.
FCRB labels follow the official BFCL checker and were kept unchanged.
To bound the effect of these labels, we run a worst-case deletion (Table~\ref{tab:label-noise}).
For each judge and dataset, we remove the share of wrong values that the audit could not confirm, 3\% on CFB and 4\% on FinTrace, choosing the values whose removal lowers the gain of \method{} over Plain the most.
This targets the way such noise favors \method{}: a reference that names the gold value leads the judge to reject a value that is labeled wrong but also correct, which counts as a correct rejection.
On both datasets, every judge keeps a significant gain, between 11.76 and 20.19 points on CFB and between 0.89 and 6.42 points on FinTrace.
We do not apply this test to the comparison with RevisEval, since its revised call also names the gold value, so the two methods share this effect and a deletion chosen against one of them would not model label noise.
The derivation gap and $\tau$ use only correct arguments, so these negative-label ambiguities do not affect Table~\ref{tab:gap}.

\begin{table}[!htbp]
  \centering
  \footnotesize
  \setlength{\tabcolsep}{3pt}
  \begin{tabular}{lcccc}
    \toprule
    & \multicolumn{2}{c}{CFB} & \multicolumn{2}{c}{FinTrace} \\
    \cmidrule(lr){2-3}\cmidrule(lr){4-5}
    Judge & Full & Worst case & Full & Worst case \\
    \midrule
    CLM & 22.14 & \makecell{20.19\\[-1pt]{\scriptsize[18.13, 22.42]}} & 5.74 & \makecell{1.87\\[-1pt]{\scriptsize[0.64, 2.90]}} \\
    Jev & 13.72 & \makecell{11.76\\[-1pt]{\scriptsize[9.79, 13.71]}} & 3.99 & \makecell{0.89\\[-1pt]{\scriptsize[0.23, 1.30]}} \\
    Qwen3-8B & 16.34 & \makecell{14.05\\[-1pt]{\scriptsize[12.82, 16.28]}} & 7.13 & \makecell{3.37\\[-1pt]{\scriptsize[2.46, 4.42]}} \\
    Qwen3.8-27B & 16.17 & \makecell{13.84\\[-1pt]{\scriptsize[11.72, 16.01]}} & 2.94 & \makecell{1.15\\[-1pt]{\scriptsize[0.42, 2.13]}} \\
    Llama-3.1-8B & 16.98 & \makecell{15.37\\[-1pt]{\scriptsize[13.80, 16.47]}} & 8.78 & \makecell{5.22\\[-1pt]{\scriptsize[4.29, 6.45]}} \\
    Mistral-7B & 20.84 & \makecell{19.70\\[-1pt]{\scriptsize[17.49, 21.99]}} & 9.16 & \makecell{6.42\\[-1pt]{\scriptsize[5.40, 7.37]}} \\
    \bottomrule
  \end{tabular}
  \caption{Worst-case deletion of unconfirmed wrong labels. Full is the gain of \method{} over Plain in argument AUROC (points) on all wrong values. Worst case removes, for each judge, the 3\% of CFB wrong values and 4\% of FinTrace wrong values whose removal lowers that gain the most, and reports the gain that remains, with its 95\% CI below. FCRB labels follow the BFCL checker and are not audited.}
  \label{tab:label-noise}
\end{table}

\paragraph{Judge input.}
Every judge reads the same text (Figure~\ref{fig:input}).
The state gives the request and the most recent earlier calls with their returns.
CFB shows the last 3 steps and FinTrace the last 5 calls of earlier turns.
Each return is cut to its first 1,000 characters.
If the longest CLM input of a state is still above 2,048 tokens, we drop the returns of the earliest shown step, then whole early steps, and last the middle of the request, keeping at least 40 tokens of it.
A state that is still too long is left out.
Correct and wrong calls at a state share the state text.
After the state come the function line, the parameter line, and the proposed value.
On FCRB the parameter line adds the type, the enum, and the default of the schema.
On CFB it adds the enum, and on FinTrace it adds nothing.
The question to the judge is ``Is this the correct value for this parameter, given the question?''
CLM is CLM-v0.1-8B.
We call Jev through OpenRouter with the version id \texttt{typesafe/jev-1.13-20260917}, and we check that every reply returns this id.
When we ask Jev again on 500 random inputs, 497 answers fall on the same side of 0.5 as before, and scores differ by at most 0.17.

\paragraph{Reference block and extra truncation.}
The reference goes after the state under the header ``Expected value (written before seeing the proposed value):''.
If the state with this block exceeds 2,048 CLM tokens, the state takes further steps of the same ladder from its own level.
This removes more early history for 131 of the 2,638 CFB arguments (39 states) and 786 of the 10,224 FinTrace arguments (194 states).
All four judges read the same \method{} input.

\paragraph{Full history.}
The full-history input shows every earlier call and return, with no window and no cut.
On FinTrace, 248 of the 10,224 arguments have an input longer than 32,768 tokens for a Qwen judge.
For these we drop the oldest returns, and then the oldest steps if needed, until the input fits.
No CFB input needs this.
\method{} + full adds the reference to this input and uses the same rule.
Jev accepts less text (Table~\ref{tab:impl}).
For 104 FinTrace arguments with full history and 102 with \method{} + full, its input drops one to three more of the oldest returns by the same rule.
Every other Jev input equals the Qwen input.
Without these arguments, no difference that involves the full history changes sign, and AUROC differences move by at most 0.46 points.

\paragraph{FCRB-Relocate.}
This set mirrors how history-only values arise in CFB: for 263 of its 279 correct history-only values (94.3\%), the source return is in the judge input but cut before the value.
We start from FCRB correct values that appear in the request.
We move one such value into the return of an added earlier call, \texttt{find\_saved\_items}, a name that no FCRB function uses.
The request is rewritten to point to the right entry by an attribute that only this entry has.
The return lists eight entries.
One holds the correct value.
Seven distractors hold values of the same kind from other FCRB samples, taken first from the same parameter name, then from the same function, then from any parameter.
In the visible version the correct entry comes second.
In the buried version it comes last.
When the correct entry would still start before character 1,000, every entry in both versions gets the same filler note, repeated until the correct entry of the buried version starts after that point, so the two versions hold the same eight entries in a different order.
In the buried returns, the correct value starts between characters 1,009 and 1,201.
The first distractor (H) is in view in both versions and starts by character 33.
The wrong values are H for every state and, where one exists, a value that appears in no text of the state (C).
This gives 257 correct values and 428 wrong values (257 H and 171 C) per version.
We drop states in which a distractor equals the correct value after lower-casing and removing all characters other than letters and digits.
Code asserts every placement above.
On the buried version, AUROC of the correct value against H alone is 34.21 for CLM, 28.23 for Jev, 26.71 for Qwen3-8B, and 39.88 for Qwen3.8-27B.
With the full history, the buried version reaches 99.81 [99.65, 99.93] for Jev, 98.39 [97.54, 99.12] for Qwen3-8B, and 99.86 [99.65, 99.99] for Qwen3.8-27B, as high as \method{} (Table~\ref{tab:controlled}).
So this set shows that judges fail when the source is out of view, not that \method{} beats the full history.
In a check of 19 states, the rewritten request and the full return fix the correct value in all 19, and H is wrong in all 19.
Two of the 19 rewrites have wording problems that do not change the label.

\paragraph{FinTrace-HC.}
We start from the 1,099 FinTrace wrong calls whose original values all appear in the request.
For the changed argument, H takes another value that the same key holds in an earlier call or return of the trajectory.
H must have the same JSON type as the correct value, must differ from it, and must not appear in the request.
Synonyms of the reporting period do not count as different.
H keeps the rest of the call and the state.
C is the rule-made wrong call of the same state, kept when all its changed values are on the \textsc{No-source} path.
We drop H values that the golden trajectory uses in a later call (35 of 172), and tickers of the same company as the correct ticker (15).
This leaves 122 H values, 72 history-only and 50 visible.
Table~\ref{tab:controlled} compares the 122 correct values with them.
The paired H minus C test of Section~\ref{sec:follow} uses the 106 of these states that also have a C (51 queries).
In a check of 20 samples, all 20 changed values are wrong at their step.

\begin{table}[!htbp]
  \centering
  \footnotesize
  \begin{tabular}{lccc}
    \toprule
    Judge & Plain & \method{} & $\Delta$ \\
    \midrule
    \multicolumn{4}{l}{\textit{Relocate, buried} (257 correct, 428 wrong)} \\
    CLM & 43.73 & 98.40 & \makecell{+54.67\\[-1pt]{\scriptsize[51.22, 58.04]}} \\
    Jev & 45.00 & 98.70 & \makecell{+53.70\\[-1pt]{\scriptsize[49.74, 57.47]}} \\
    Qwen3-8B & 39.80 & 98.87 & \makecell{+59.07\\[-1pt]{\scriptsize[55.35, 62.76]}} \\
    Qwen3.8-27B & 49.03 & 99.59 & \makecell{+50.56\\[-1pt]{\scriptsize[46.98, 53.92]}} \\
    \midrule
    \multicolumn{4}{l}{\textit{Relocate, visible} (257 correct, 428 wrong)} \\
    CLM & 91.89 & 99.42 & \makecell{+7.52\\[-1pt]{\scriptsize[5.64, 9.53]}} \\
    Jev & 99.88 & 99.99 & \makecell{+0.11\\[-1pt]{\scriptsize[0.03, 0.21]}} \\
    Qwen3-8B & 98.82 & 99.68 & \makecell{+0.87\\[-1pt]{\scriptsize[0.17, 1.64]}} \\
    Qwen3.8-27B & 99.58 & 99.95 & \makecell{+0.37\\[-1pt]{\scriptsize[0.06, 0.81]}} \\
    \midrule
    \multicolumn{4}{l}{\textit{FinTrace-HC} (122 correct, 122 wrong)} \\
    CLM & 91.98 & 96.45 & \makecell{+4.46\\[-1pt]{\scriptsize[1.49, 7.76]}} \\
    Jev & 98.90 & 99.32 & \makecell{+0.42\\[-1pt]{\scriptsize[$-$0.38, 1.25]}} \\
    Qwen3-8B & 95.99 & 97.95 & \makecell{+1.97\\[-1pt]{\scriptsize[0.15, 4.28]}} \\
    Qwen3.8-27B & 98.56 & 99.26 & \makecell{+0.70\\[-1pt]{\scriptsize[$-$0.04, 1.55]}} \\
    \bottomrule
  \end{tabular}
  \caption{Argument AUROC (in points) on the two controlled sets. $\Delta$ is \method{} minus Plain, with its 95\% CI below. No difference is significantly negative.}
  \label{tab:controlled}
\end{table}

\paragraph{Revealing the source.}
The test uses FinTrace correct history-only values whose first occurrence was lost from the window.
We exclude booleans, empty strings, and numbers of at most two characters.
We also exclude other numbers and strings of at most three characters, which can match by chance.
We keep the 167 values whose source step lies outside the window (130 states, 93 queries).
The reveal version inserts that step at its place in time.
A long return shows 1,000 characters centered on the value.
The placebo inserts another step of the same kind from outside the window of the same trajectory.
It does not contain the value and has the closest displayed length.
Both versions use the same header.
The revealed step is a return for 80 values, an earlier call of the agent for 82, and assistant text for 5.
Against the placebo, the source raises the mean score of correct values by between 6.30 and 14.29 points for all four judges, while wrong values in the same states move by at most 3.49 points.
On revealed returns alone, the source raises correct scores significantly for all four judges.

\paragraph{Schema facts.}
For Figure~\ref{fig:follow}b we add one line after the proposed value on every FCRB argument whose schema has a default (1,037 items).
The line restates the default.
It then quotes up to 40 characters around each check word found in the request, or lists the check words when none is found.
The check words are the parameter name, the name with spaces for underscores, the default value, and the enum values.

\section{Derivation Path Census}
\label{app:census}
We assign paths in this order, and the first match wins.

\textsc{Visible}: the value is found in the content of the Plain state, which is the request, the shown calls, and the shown returns as cut.
Template text, the function and parameter lines, and the proposed value do not count.
\textsc{History-only}: the value is found in the full history, which is the request and every earlier call and return, uncut.
\textsc{Format}: the value matches the request or a piece of the full history after one change.
The change can be lower-casing and collapsing whitespace, or reading both sides as the same calendar date, the same number, or the same boolean.
\textsc{Default}: the value equals the schema default, or it is a ``don't care'' value that the parameter description names.
\textsc{No-source}: none of the above.
A value is found when its text is a case-sensitive substring of the search text.
A number must not touch a letter, a digit, or a decimal point, so \texttt{10} does not match \texttt{2010} or \texttt{10.5}.
A string shorter than four characters must not touch a letter or a digit.
A list is found when every element is found.
The path depends only on the value under test, so wrong values have paths too.
Only FCRB schemas have defaults, so the \textsc{Default} path occurs only on FCRB.

\section{Metrics and Statistics}
\label{app:metrics}
\paragraph{Threshold and ties.}
For each judge, dataset, and input, $\tau$ is the score at which the judge accepts 80\% of correct arguments.
Scores above $\tau$ count as accepted.
Scores equal to $\tau$ count with a fraction $q$ chosen so that exactly 80\% are accepted.
This matters for judges whose scores pile up at one value.
The choice of 80\% does not drive Table~\ref{tab:gap}.
Table~\ref{tab:tau-sens} repeats its six columns with $\tau$ set to accept 70\% and 90\% of correct arguments, for the five judges run at all three rates.
On CFB and FinTrace, all 20 cells keep a CI above zero at every rate.
On FCRB, where the gap is small, three of the 30 cells lose significance at 70\% or 90\%.

\begin{table*}[t]
\centering
\small
\setlength{\tabcolsep}{4pt}
\begin{tabular}{llcccccc}
\toprule
& & \multicolumn{3}{c}{Out of view} & \multicolumn{3}{c}{Main path} \\
\cmidrule(lr){3-5} \cmidrule(lr){6-8}
Dataset & Judge & 70\% & 80\% & 90\% & 70\% & 80\% & 90\% \\
\midrule
    FCRB (Default) & CLM & \makecell{5.85\\[-1pt]{\scriptsize[1.30, 10.89]}} & \makecell{8.15\\[-1pt]{\scriptsize[3.92, 12.07]}} & \makecell{6.20\\[-1pt]{\scriptsize[3.58, 9.51]}} & \makecell{32.60\\[-1pt]{\scriptsize[24.82, 41.14]}} & \makecell{31.93\\[-1pt]{\scriptsize[24.36, 41.20]}} & \makecell{26.29\\[-1pt]{\scriptsize[19.87, 34.75]}} \\
     & Jev & \makecell{11.82\\[-1pt]{\scriptsize[7.35, 15.90]}} & \makecell{8.11\\[-1pt]{\scriptsize[4.44, 11.74]}} & \makecell{2.22\\[-1pt]{\scriptsize[$-$0.75, 4.77]}} & \makecell{32.99\\[-1pt]{\scriptsize[24.41, 41.88]}} & \makecell{20.26\\[-1pt]{\scriptsize[12.68, 27.32]}} & \makecell{9.41\\[-1pt]{\scriptsize[3.16, 14.70]}} \\
     & Qwen3-4B & \makecell{3.77\\[-1pt]{\scriptsize[$-$0.43, 7.78]}} & \makecell{4.37\\[-1pt]{\scriptsize[0.37, 8.43]}} & \makecell{3.72\\[-1pt]{\scriptsize[0.75, 6.50]}} & \makecell{25.82\\[-1pt]{\scriptsize[18.43, 34.44]}} & \makecell{21.26\\[-1pt]{\scriptsize[13.91, 29.63]}} & \makecell{15.64\\[-1pt]{\scriptsize[8.63, 22.33]}} \\
     & Qwen3-8B & \makecell{1.15\\[-1pt]{\scriptsize[$-$2.98, 4.88]}} & \makecell{0.63\\[-1pt]{\scriptsize[$-$3.78, 4.35]}} & \makecell{$-$1.00\\[-1pt]{\scriptsize[$-$3.85, 1.77]}} & \makecell{11.97\\[-1pt]{\scriptsize[3.59, 19.85]}} & \makecell{14.40\\[-1pt]{\scriptsize[4.69, 22.14]}} & \makecell{3.52\\[-1pt]{\scriptsize[$-$1.60, 9.31]}} \\
     & Qwen3.8-27B & \makecell{16.93\\[-1pt]{\scriptsize[11.92, 21.19]}} & \makecell{12.29\\[-1pt]{\scriptsize[8.19, 15.81]}} & \makecell{6.07\\[-1pt]{\scriptsize[3.26, 8.78]}} & \makecell{49.55\\[-1pt]{\scriptsize[40.32, 56.73]}} & \makecell{36.03\\[-1pt]{\scriptsize[27.56, 44.48]}} & \makecell{16.88\\[-1pt]{\scriptsize[10.16, 23.69]}} \\
    \midrule
    CFB (History only) & CLM & \makecell{41.68\\[-1pt]{\scriptsize[38.43, 45.44]}} & \makecell{30.76\\[-1pt]{\scriptsize[27.95, 34.02]}} & \makecell{17.15\\[-1pt]{\scriptsize[15.33, 18.97]}} & \makecell{77.22\\[-1pt]{\scriptsize[70.77, 82.83]}} & \makecell{66.63\\[-1pt]{\scriptsize[59.42, 73.00]}} & \makecell{38.19\\[-1pt]{\scriptsize[32.88, 44.93]}} \\
     & Jev & \makecell{28.50\\[-1pt]{\scriptsize[22.35, 33.74]}} & \makecell{22.67\\[-1pt]{\scriptsize[19.08, 27.71]}} & \makecell{16.56\\[-1pt]{\scriptsize[13.77, 18.55]}} & \makecell{74.59\\[-1pt]{\scriptsize[66.98, 79.65]}} & \makecell{59.39\\[-1pt]{\scriptsize[54.22, 68.00]}} & \makecell{42.09\\[-1pt]{\scriptsize[35.21, 49.18]}} \\
     & Qwen3-4B & \makecell{47.34\\[-1pt]{\scriptsize[44.29, 50.65]}} & \makecell{34.71\\[-1pt]{\scriptsize[32.87, 36.78]}} & \makecell{18.86\\[-1pt]{\scriptsize[17.73, 20.08]}} & \makecell{86.84\\[-1pt]{\scriptsize[82.10, 91.14]}} & \makecell{76.81\\[-1pt]{\scriptsize[71.30, 81.93]}} & \makecell{50.29\\[-1pt]{\scriptsize[44.37, 57.15]}} \\
     & Qwen3-8B & \makecell{43.72\\[-1pt]{\scriptsize[40.11, 47.33]}} & \makecell{33.30\\[-1pt]{\scriptsize[31.13, 35.65]}} & \makecell{18.96\\[-1pt]{\scriptsize[17.95, 20.33]}} & \makecell{83.63\\[-1pt]{\scriptsize[78.15, 88.16]}} & \makecell{74.87\\[-1pt]{\scriptsize[68.09, 80.47]}} & \makecell{46.56\\[-1pt]{\scriptsize[40.84, 53.62]}} \\
     & Qwen3.8-27B & \makecell{34.79\\[-1pt]{\scriptsize[28.64, 40.36]}} & \makecell{25.99\\[-1pt]{\scriptsize[21.10, 30.72]}} & \makecell{14.47\\[-1pt]{\scriptsize[11.85, 17.66]}} & \makecell{76.27\\[-1pt]{\scriptsize[69.90, 82.04]}} & \makecell{57.08\\[-1pt]{\scriptsize[50.28, 63.87]}} & \makecell{28.22\\[-1pt]{\scriptsize[23.28, 36.22]}} \\
    \midrule
    FinTrace (No source) & CLM & \makecell{32.99\\[-1pt]{\scriptsize[30.19, 36.14]}} & \makecell{25.95\\[-1pt]{\scriptsize[23.46, 28.22]}} & \makecell{15.19\\[-1pt]{\scriptsize[13.79, 16.78]}} & \makecell{34.67\\[-1pt]{\scriptsize[31.45, 38.15]}} & \makecell{28.28\\[-1pt]{\scriptsize[25.34, 30.81]}} & \makecell{17.09\\[-1pt]{\scriptsize[15.48, 18.93]}} \\
     & Jev & \makecell{36.44\\[-1pt]{\scriptsize[33.57, 39.11]}} & \makecell{27.78\\[-1pt]{\scriptsize[25.20, 29.87]}} & \makecell{15.79\\[-1pt]{\scriptsize[14.20, 17.37]}} & \makecell{36.06\\[-1pt]{\scriptsize[32.88, 39.00]}} & \makecell{27.47\\[-1pt]{\scriptsize[24.54, 30.25]}} & \makecell{16.53\\[-1pt]{\scriptsize[14.56, 18.45]}} \\
     & Qwen3-4B & \makecell{30.54\\[-1pt]{\scriptsize[27.77, 33.99]}} & \makecell{23.30\\[-1pt]{\scriptsize[20.91, 25.94]}} & \makecell{13.52\\[-1pt]{\scriptsize[11.86, 15.30]}} & \makecell{30.44\\[-1pt]{\scriptsize[27.39, 34.24]}} & \makecell{23.29\\[-1pt]{\scriptsize[20.73, 26.35]}} & \makecell{13.79\\[-1pt]{\scriptsize[11.80, 15.84]}} \\
     & Qwen3-8B & \makecell{31.48\\[-1pt]{\scriptsize[28.91, 34.47]}} & \makecell{24.04\\[-1pt]{\scriptsize[22.13, 26.22]}} & \makecell{14.42\\[-1pt]{\scriptsize[12.97, 15.81]}} & \makecell{32.93\\[-1pt]{\scriptsize[30.18, 36.21]}} & \makecell{25.72\\[-1pt]{\scriptsize[23.59, 28.31]}} & \makecell{15.80\\[-1pt]{\scriptsize[14.02, 17.53]}} \\
     & Qwen3.8-27B & \makecell{40.02\\[-1pt]{\scriptsize[37.07, 42.60]}} & \makecell{29.76\\[-1pt]{\scriptsize[27.44, 32.10]}} & \makecell{16.47\\[-1pt]{\scriptsize[14.98, 17.83]}} & \makecell{42.28\\[-1pt]{\scriptsize[39.03, 45.30]}} & \makecell{31.68\\[-1pt]{\scriptsize[28.98, 34.22]}} & \makecell{17.65\\[-1pt]{\scriptsize[15.97, 19.38]}} \\
\bottomrule
\end{tabular}
\caption{Derivation gap $\mathrm{Gap}_\tau$ (in points) when $\tau$ is set to accept 70\%, 80\%, or 90\% of correct arguments, with Plain input. The 80\% columns repeat Table~\ref{tab:gap}. Out of view pools the four paths other than \textsc{Visible}; the main path of each dataset is in parentheses. Llama-3.1-8B and Mistral-7B were run at 80\% only, as additional rows of Table~\ref{tab:gap}, so the check covers the five judges run at all three rates. 95\% CIs come from the same bootstrap, with $\tau$ and $q$ set again in every resample.}
\label{tab:tau-sens}
\end{table*}

\paragraph{Bootstrap.}
We resample queries 1,000 times with seed 20260930.
For argument AUROC we draw queries with replacement.
For call AUROC and $\mathrm{Gap}_\tau$ we give the queries multinomial weights, which has the same distribution but a different random sequence.
For $\mathrm{Gap}_\tau$, $\tau$ and $q$ are set again in each bootstrap sample.
A 95\% CI runs from the 2.5th to the 97.5th percentile over these samples.
A paired difference computes both sides on the same samples.

\begin{table*}[!htbp]
  \centering
  \footnotesize
  \begin{tabular}{llrccc}
    \toprule
    & Value & Correct / wrong & \method{} & EV only & $\Delta$ \\
    \midrule
    FCRB & Stated & 1,577 / 620 & 94.69 & 94.93 & \makecell{$-$0.24\\[-1pt]{\scriptsize[$-$0.89, 0.40]}} \\
     & Not found & 142 / 149 & 72.37 & 73.09 & \makecell{$-$0.72\\[-1pt]{\scriptsize[$-$4.24, 3.13]}} \\
    \midrule
    CFB & Stated & 1,290 / 523 & 97.78 & 96.39 & \makecell{+1.39\\[-1pt]{\scriptsize[0.85, 2.05]}} \\
     & Not found & 348 / 470 & 82.78 & 60.41 & \makecell{+22.37\\[-1pt]{\scriptsize[16.99, 27.46]}} \\
    \midrule
    FinTrace & Stated & 3,384 / 1,763 & 95.99 & 97.11 & \makecell{$-$1.13\\[-1pt]{\scriptsize[$-$1.46, $-$0.75]}} \\
     & Not found & 4,044 / 1,033 & 82.50 & 78.83 & \makecell{+3.67\\[-1pt]{\scriptsize[2.26, 5.33]}} \\
    \bottomrule
  \end{tabular}
  \caption{Keeping only the parsed expected value (EV only), split by whether the parser finds a value stated in the reference of the argument slot. Argument AUROC (in points) with Qwen3.8-27B. $\Delta$ is \method{} minus EV only, with its 95\% CI below. On CFB, the seven arguments of the one deriver prompt without an answer are left out.}
  \label{tab:parser-e}
\end{table*}
\section{Reading the Reference}
\label{app:parser}
Some analyses need the value that a reference states.
They are the ablation that keeps only the expected value, the swap test, and the shares of references that name the correct value (Section~\ref{sec:paths}).
A rule-based parser reads this value.
It first removes Markdown emphasis.
It then matches a fixed set of sentence patterns, such as ``should be'' followed by a value, and takes the value as a substring of the reference.
If no pattern matches, the reference counts as not fixing the value.
It also counts as not fixing the value when the same sentence gives a second candidate value.
This last rule changes 1, 22, and 51 argument slots on FCRB, CFB, and FinTrace.
Parallel calls to one function at a step share a slot, and the reference then lists the values it expects, so these slots fall under this rule.

\paragraph{Checks.}
We check the parser on samples drawn with seed 20260930, and no argument is drawn twice.
For each benchmark, a check reads 20 correct and 10 wrong arguments for which the parser finds a value, and asks whether that value is the conclusion of the reference.
We fixed the parser after each failed round.
In the fourth round, it took the right value for 30 of 30 arguments on every benchmark, so we accepted it.
The parser can also miss a stated value.
So the first two rounds also read 20 correct arguments per benchmark for which the parser found no value although the reference text contains the correct value.
In the second round, the reference does state a value for some of them, so the parser misses a small share of stated values.
The last two rounds changed only one fallback pattern and added the rule for a second candidate value.
A miss counts a stated value as not fixed, so the shares of references that name the correct value can only be too low.

\paragraph{Parsed expected value only.}
This ablation replaces the text of each reference block by one sentence.
When the parser finds a value, the sentence is ``Expected value for \{name\}: \{value\}.'', with the value copied from the reference.
Otherwise it is ``The context does not determine the value of \{name\}.''
Everything else is the same as in \method{}, including the header of the block and its place in the input.
Table~\ref{tab:parser-e} splits the result by whether the parser finds a value for the argument slot.
Misses of the parser fall only into the second group, so the first group does not depend on them.
Where a value is stated, the rest of the text adds 1.39 points on CFB, makes no significant difference on FCRB, and costs 1.13 points on FinTrace.
Where no value is found, it adds 22.37 points on CFB and 3.67 points on FinTrace.
There the reference text still contains the correct value for 77.01\% and 45.03\% of the correct arguments.

\begin{table*}[!htbp]
\centering
\footnotesize
\setlength{\tabcolsep}{4pt}
\begin{tabular}{lccccccccc}
  \toprule
  & \multicolumn{3}{c}{FCRB} & \multicolumn{3}{c}{CFB} & \multicolumn{3}{c}{FinTrace} \\
  \cmidrule(lr){2-4}\cmidrule(lr){5-7}\cmidrule(lr){8-10}
  Judge & Plain & \method{} & $\Delta$ & Plain & \method{} & $\Delta$ & Plain & \method{} & $\Delta$ \\
  \midrule
  CLM & 66.19 & 85.84 & \makecell{+19.66\\[-1pt]{\scriptsize[17.10, 22.27]}} & 52.56 & 73.47 & \makecell{+20.91\\[-1pt]{\scriptsize[17.38, 24.76]}} & 80.34 & 81.55 & \makecell{+1.21\\[-1pt]{\scriptsize[0.00, 2.37]}} \\
  Jev & 83.46 & 88.12 & \makecell{+4.67\\[-1pt]{\scriptsize[3.15, 6.11]}} & 56.56 & 78.67 & \makecell{+22.11\\[-1pt]{\scriptsize[18.18, 25.96]}} & 87.50 & 90.06 & \makecell{+2.56\\[-1pt]{\scriptsize[1.88, 3.25]}} \\
  Qwen3-8B & 73.98 & 86.07 & \makecell{+12.09\\[-1pt]{\scriptsize[9.80, 14.19]}} & 55.47 & 67.46 & \makecell{+11.99\\[-1pt]{\scriptsize[8.62, 15.51]}} & 82.19 & 87.31 & \makecell{+5.12\\[-1pt]{\scriptsize[4.08, 6.21]}} \\
  Qwen3.8-27B & 85.43 & 88.72 & \makecell{+3.29\\[-1pt]{\scriptsize[1.74, 4.83]}} & 68.78 & 85.98 & \makecell{+17.20\\[-1pt]{\scriptsize[13.84, 20.55]}} & 86.69 & 88.08 & \makecell{+1.39\\[-1pt]{\scriptsize[0.63, 2.20]}} \\
  \bottomrule
\end{tabular}
\caption{Call AUROC (in points) with Plain input and with \method{}. $\Delta$ is \method{} minus Plain, with its 95\% CI below. Only the argument scores change with the input. For CLM on FinTrace the lower bound is 0.0005 points.}
\label{tab:grid-call}
\end{table*}

\section{Call-Level Results}
\label{app:call}
Guards accept or reject a whole call \citep{she2026safeguarding}, so we also score each call from its argument scores.
Process reward models combine step scores by a product or a minimum \citep{lightman2024let,zhang2025lessons}, and the call score takes the minimum.
\paragraph{Call score.}
For a call with argument scores $s_i$ and whole-call score $w$, the call score is
\begin{equation}
  s_{\mathrm{call}} = \min\Big(w,\; \min_i s_i\Big),
\end{equation}
The whole-call score always comes from the Plain input.
Only the argument scores change with the input.
Most arguments of a wrong call are not the changed one and carry no label.
All but 134 of them have the same judge input, word for word, as an argument of the correct call at the same state, and they take that argument's score under every input.
The other 134, all on FCRB, get their own score under every input.

\paragraph{Decomposed verification.}
Fact checkers often split a text into atomic claims and verify each one \citep{min2023factscore}.
Atomic claims can lose the context needed to check them \citep{gunjal2024molecular,wanner2025dndscore}, and splitting helps weaker verifiers and can hurt stronger ones \citep{hu2025decomposition}.
Process reward models also combine step scores, and the best rule depends on how the step labels were made \citep{lightman2024let,zhang2025lessons}.
An argument judge checks a call in the same way, one argument at a time.

\paragraph{The gap costs at the call level.}
Taking the minimum assumes that the weakest argument of a call is a wrong one.
A correct value out of view often breaks this assumption (Table~\ref{tab:min-arg}).
Compared with the whole-call score alone, the minimum raises call AUROC on FCRB, the dataset with the smallest gap, for every judge by between 8.15 and 16.21 points.
On CFB, which has the largest gap on a single path, it adds nothing for any judge (from $-3.04$ to $1.22$ points, none significant).

\begin{table}[!htbp]
\centering
\footnotesize
\setlength{\tabcolsep}{3pt}
\begin{tabular}{lccc}
  \toprule
  & FCRB & CFB & FinTrace \\
  \midrule
  Mixed correct calls & 325 & 197 & 1,107 \\
  Chance & 50.45 & 48.77 & 48.73 \\
  \midrule
  CLM & \makecell{62.77\\[-1pt]{\scriptsize[7.33, 17.62]}} & \makecell{93.40\\[-1pt]{\scriptsize[40.97, 48.01]}} & \makecell{75.07\\[-1pt]{\scriptsize[23.28, 29.28]}} \\
  Jev & \makecell{62.72\\[-1pt]{\scriptsize[7.49, 16.93]}} & \makecell{85.25\\[-1pt]{\scriptsize[32.34, 40.93]}} & \makecell{84.19\\[-1pt]{\scriptsize[32.85, 38.15]}} \\
  Qwen3-4B & \makecell{56.02\\[-1pt]{\scriptsize[2.02, 9.23]}} & \makecell{91.53\\[-1pt]{\scriptsize[38.63, 46.33]}} & \makecell{83.59\\[-1pt]{\scriptsize[32.23, 37.36]}} \\
  Qwen3-8B & \makecell{48.96\\[-1pt]{\scriptsize[$-$5.02, 2.11]}} & \makecell{90.36\\[-1pt]{\scriptsize[37.99, 45.27]}} & \makecell{82.11\\[-1pt]{\scriptsize[30.98, 35.66]}} \\
  Qwen3.8-27B & \makecell{68.31\\[-1pt]{\scriptsize[12.89, 22.59]}} & \makecell{82.49\\[-1pt]{\scriptsize[28.37, 39.00]}} & \makecell{85.68\\[-1pt]{\scriptsize[34.68, 39.40]}} \\
  Llama-3.1-8B & \makecell{63.33\\[-1pt]{\scriptsize[8.38, 17.17]}} & \makecell{70.05\\[-1pt]{\scriptsize[14.98, 27.73]}} & \makecell{79.27\\[-1pt]{\scriptsize[27.52, 33.15]}} \\
  Mistral-7B & \makecell{59.89\\[-1pt]{\scriptsize[5.57, 13.40]}} & \makecell{94.62\\[-1pt]{\scriptsize[42.31, 49.05]}} & \makecell{82.01\\[-1pt]{\scriptsize[30.73, 35.89]}} \\
  \bottomrule
\end{tabular}
\caption{Share (\%) of correct calls whose lowest-scored argument is out of view, among correct calls with arguments both in and out of view, under Plain input. Chance is the share of out-of-view arguments in these calls. Below each share is the 95\% CI of its difference from chance, in points. No share is significantly below chance.}
\label{tab:min-arg}
\end{table}

\paragraph{The lowest argument of a correct call.}
Table~\ref{tab:min-arg} shows which argument takes the minimum in correct calls with arguments both in and out of view.
For 20 of the 21 pairs of judge and benchmark, the lowest-scored argument is out of view more often than chance, and for none less often.
The one exception is Qwen3-8B on FCRB, where about half of these calls tie for the lowest score.
On CFB, \method{} lowers this share for all four judges, by between 10.28 and 24.37 percentage points, but it stays above chance.

\paragraph{\method{} at the call level.}
Table~\ref{tab:grid-call} gives call AUROC with Plain and with \method{}.
\method{} raises call AUROC for all four judges on all three benchmarks, by between 1.21 and 22.11 points, and all 12 gains are significant.
On CFB, where taking the minimum adds nothing under Plain, \method{} raises call AUROC by between 11.99 and 22.11 points.

\begin{table*}[!htbp]
\centering
\footnotesize
\begin{tabular}{p{0.17\textwidth}p{0.77\textwidth}}
  \toprule
  Item & Setting \\
  \midrule
  CLM & CLM-v0.1-8B \citep{kwok2026contrastive}, one checkpoint file (SHA-256 b2b4a8c9\ldots). We serve it with vLLM 0.30.0 in bf16. The score is the probability it gives to the answer ``true'' against ``false''. We run it with the released serving settings, which truncate inputs to 2,048 tokens. \\
  Jev & \texttt{typesafe/jev-1.13-20260917} \citep{almeida2026introducing}, called through OpenRouter. In a length probe, it read inputs of up to 32,831 tokens by its own count and refused longer ones. A request holds only the state, the model id, and one question. We use the returned probability and set no sampling parameters. \\
  Qwen judges & Qwen3-4B and 8B \citep{yang2025qwen3}, and Qwen3.8-27B \citep{qwenteam2026qwen3827b}, with 4.02B, 8.19B, and 27.78B parameters (26.90B in the text part of Qwen3.8-27B). We use one revision of each (1cfa9a7, b968826, and 1d4bf0f). They run in bf16 with transformers 5.18.0 and left padding. Thinking is off. The score is the softmax over the logits of ``Yes'' and ``No'' at the first output position. Where a judge runs with thinking on (Tables~\ref{tab:reference-thinking} and~\ref{tab:deriver-judge}), it first generates its thinking, up to 8,192 tokens, after which we append the closing tag if it has not been written, and we then read the same two logits at the first position of the answer, without parsing the answer text. The two tokens hold less than half of the probability mass for at most 1.92\% of inputs on any benchmark. Truncated inputs run in batches of 32. Full-history inputs run in batches sorted by length, at most 32 inputs and at most 65,536 padded tokens for Qwen3-8B and 32,768 for Qwen3.8-27B. \\
  Other LLM judges & Llama-3.1-8B-Instruct \citep{grattafiori2024llama3} and Mistral-7B-Instruct-v0.3 \citep{jiang2023mistral}, revisions 0e9e39f and c170c70. We read them as the Qwen judges, in bf16 with left padding, with the same system prompt and question. Neither tokenizer has a padding token, so we pad with the end-of-sequence token. \\
  Generation & Qwen3.8-27B with vLLM 0.30.0 in bf16, with thinking on at medium reasoning effort. Deriver prompts hold up to 32,768 tokens, and the oldest returns are dropped beyond that, which affects 190 of the 7,428 FinTrace deriver prompts. The deriver samples once per prompt with temperature 1.0, top-p 0.95, and top-k 20, the values recommended in the generation config of the model, and with seed 20260930. It writes at most 8,192 new tokens, and the reference is the text after the thinking. If no text follows the thinking, the judge gets the Plain input. This happens for one CFB prompt, which serves seven arguments, and no deriver prompt reaches the limit. ProvenanceGuard evidence and RevisEval use the same model and decoding settings. In the ablation, Qwen3-8B as deriver uses the values recommended for it, with temperature 0.6, and the deriver without thinking uses greedy decoding and at most 1,024 new tokens. \\
  Hardware, software & Every job runs on one NVIDIA B200 GPU. Models run with Python 3.11.16, PyTorch 2.13.0 with CUDA 13.0, vLLM 0.30.0 \citep{kwon2023efficient}, and transformers 5.18.0. Analysis uses Python 3.12.11, NumPy 2.2.6, pandas 2.3.2, and SciPy 1.16.1. \\
  Statistics & AUROC uses our own rank code, which counts ties as one half. CIs come from the bootstrap in Appendix~\ref{app:metrics}. \\
  Randomness & Seed 20260930 fixes every random step: the sampling of the deriver and the prior methods, the FinTrace wrong calls, the two controlled sets, the swap arguments, the case studies, and the bootstrap. Qwen scores shift a little with batch composition. In our checks, one score moved by at most 0.185 and argument AUROC by at most 0.89 points. A rerun of CLM on all CFB arguments leaves 98.2\% of scores unchanged and AUROC the same. \\
  Compute & The results in this paper took 44 jobs and 41.98 GPU hours, with model loading: 6.31 for generation, 27.66 for judges on truncated inputs, and 8.01 for judges on the full history. Runs of variants that we do not report took another 15.40, and pilot runs, smoke tests, exploration, and failed jobs took 12.75. Jev took 325,637 billed calls for \$23.01, and 47,514 more for variants that we do not report. \\
  \bottomrule
\end{tabular}
\caption{Models and settings.}
\label{tab:impl}
\end{table*}
\begin{table*}[!htbp]
\centering
\footnotesize
\setlength{\tabcolsep}{2.5pt}
\begin{tabular}{lrrrrrrrr}
  \toprule
  & \multicolumn{5}{c}{Deriver (Qwen3.8-27B)} & \multicolumn{3}{c}{Judge input (Qwen3-8B tokens)} \\
  \cmidrule(lr){2-6}\cmidrule(lr){7-9}
  Benchmark & Prompts & Candidates / prompt & Input tokens & Output tokens & GPU s & Plain & \method{} & Full \\
  \midrule
  FCRB & 1,871 & 1.33 & 195 (365) & 217 (968) & 12 & 165 (352) & 236 (444) & 165 (352) \\
  CFB & 1,442 & 1.83 & 2,892 (29,327) & 513 (4,763) & 138 & 841 (2,031) & 1,044 (2,084) & 3,317 (29,673) \\
  FinTrace & 7,428 & 1.38 & 6,853 (32,651) & 557 (6,351) & 1,684 & 1,093 (2,084) & 1,169 (2,084) & 6,761 (32,646) \\
  \bottomrule
\end{tabular}
\caption{Cost of \method{}. Token counts give the mean and, in parentheses, the maximum. Deriver tokens use the tokenizer and chat template of Qwen3.8-27B, and judge tokens those of Qwen3-8B, so they include the template text. Output tokens include the thinking. GPU seconds cover generation without model loading. The run also held other prompts, so we split its time by prompt tokens.}
\label{tab:cost}
\end{table*}

\section{Implementation Details and Cost}
\label{app:impl}

\begin{table*}[!t]
\centering
\footnotesize
\setlength{\tabcolsep}{4pt}
\begin{tabular}{lccc}
  \toprule
  Per input, batch size 1 & FCRB & CFB & FinTrace \\
  \midrule
  CLM, \method{} input (ms) & 8.7 & 8.0 & 8.0 \\
  Qwen3-8B, \method{} input (ms) & 34.7 & 66.9 & 57.6 \\
  Qwen3.8-27B, \method{} input (ms) & 52.9 & 120.4 & 93.8 \\
  Qwen3.8-27B, direct judge with thinking, truncated input (s) & 2.00 & 7.46 & 4.52 \\
  Qwen3.8-27B, direct judge with thinking, full history (s) & -- & 11.20 & 5.53 \\
  \midrule
  Deriver, one reference prompt (s) & 2.44 & 7.27 & 6.34 \\
  \bottomrule
\end{tabular}
\caption{Wall-clock time per input on one GPU in bf16, mean over 200 inputs drawn at random from each benchmark. The deriver row is the time to write one reference, which does not depend on the candidate and is shared by every candidate of that argument at that state.}
\label{tab:latency}
\end{table*}

\paragraph{Models and settings.}
Table~\ref{tab:impl} lists the models and settings.
We train no model and search no hyperparameters.

\paragraph{Cost.}
Table~\ref{tab:cost} gives the cost of \method{}.
The deriver writes one reference per argument slot of a state, and all candidates for that slot share it.
A slot is a pair of state and parameter name, so parallel calls to one function at a step share a slot, and a parameter that only the wrong call fills gets a slot with no correct value.
One deriver prompt serves 1.33 candidates on FCRB, 1.83 on CFB, and 1.38 on FinTrace on average.
The judge reads the reference inside its usual input.
The longest \method{} input has 408 tokens on FCRB and 2,048 on CFB and FinTrace, as CLM counts them against its 2,048-token setting.
A judge that reads the full history reads a long input for every candidate, with 3,317 tokens on average on CFB and 6,761 on FinTrace.
ProvenanceGuard evidence also runs Qwen3.8-27B, but it needs one prompt per candidate value, 15,141 in all after removing prompts with identical text, against 10,741 deriver prompts.

\section{Further Results}
\label{app:more}

\paragraph{A wrong value from the history.}
FinTrace-HC pairs two wrong values for the same argument in the same state, one that appeared earlier in the history (H) and one that did not (C).
On the 106 pairs in which both values exist, the mean score of H minus C is 7.79 [5.08, 10.39] for CLM, 5.06 [3.08, 7.36] for Jev, 3.79 [0.88, 7.53] for Qwen3-8B, and 0.72 [$-$0.08, 2.01] for Qwen3.8-27B, with query-clustered bootstrap CIs; the first three are significant.

\paragraph{The reference on top of the full history.}
On CFB, \method{} with the full history reaches 84.97 for Qwen3-8B and 96.01 for Qwen3.8-27B.
It beats the full history alone by 18.04 and 10.56 points, and \method{} on the truncated input by 1.84 and 0.92 points.
Its largest gain over the full history is on no-source values, 32.87 and 18.87 points.
On FinTrace, it reaches 91.94 and 92.52, and it beats the full history alone by 7.20 and 2.75 points.
For Jev, it beats the full history alone by 9.35 points on CFB and 3.18 on FinTrace.

\paragraph{A deriver that sees the value.}
\method{}-v tests the choice of a deriver that does not see the candidate.
Its deriver gets the \method{} prompt with the candidate value added and the opening words ``Before seeing any proposed value:'' removed.
Everything else is the same, including the header of the reference block, which still says that the reference was written before seeing the proposed value.
We keep this header so that the judge input changes only through the reference itself.
Its references differ in length from those of \method{}, so 152 CFB inputs and 212 FinTrace inputs keep a different amount of early history within 2,048 tokens.

Averaged over the four judges, \method{}-v is above \method{} on CFB by 2.68 points, and below it on FCRB and FinTrace by 2.57 and 0.84 points (Table~\ref{tab:grid-main}).
It also needs one derivation per candidate value, 15,275 prompts after removing prompts with identical text, against 10,741, since candidates cannot share a reference that names one of them.
\method{}-v names the wrong value under test as the expected value for 6.61\% of wrong CFB arguments and 10.92\% of wrong FCRB arguments, against 1.10\% and 8.19\% for \method{}.
RevisEval keeps the wrong value under test in its revised call for 17.02\% and 21.46\%, and for 12.66\% of wrong FinTrace arguments, against 1.00\% for \method{}-v and 0.43\% for \method{}.
At 0.5, Qwen3.8-27B accepts 18.42\% of wrong CFB values with \method{}-v, against 8.61\% with \method{}, and 21.20\% of wrong FCRB values, against 13.00\%.
Over the three benchmarks, \method{}-v writes the wrong value under test as the expected value for 178 wrong arguments.
In a sample of 10 of them, 8 wrong values are close to the correct one, such as a long base64 string whose decoded text differs in one or two characters, a longitude with fewer digits, or a date one day off.

\paragraph{Deriver seed.}
We sample the deriver again with two more seeds and score CLM, Qwen3-8B, and Qwen3.8-27B with each new reference.
\method{} is significantly above Plain for all three seeds, all three judges, and all three benchmarks, 27 cases in all.
Against the original seed, none of the 18 differences of the two new seeds in \method{} AUROC is significant, and the largest is 0.80 points.

\paragraph{Judges outside the Qwen family.}
Table~\ref{tab:gap} includes Llama-3.1-8B-Instruct and Mistral-7B-Instruct-v0.3, and every one of their 12 gaps there is significant.
We also run \method{} with them, with the same references as for the other judges (Table~\ref{tab:other-judges}).
\method{} raises their argument AUROC on all three benchmarks, by between 8.78 and 20.84 points, and all six gains are significant.

\begin{table*}[!htbp]
\centering
\footnotesize
\setlength{\tabcolsep}{4pt}
\begin{tabular}{lccccccccc}
  \toprule
  & \multicolumn{3}{c}{FCRB} & \multicolumn{3}{c}{CFB} & \multicolumn{3}{c}{FinTrace} \\
  \cmidrule(lr){2-4}\cmidrule(lr){5-7}\cmidrule(lr){8-10}
  Judge & Plain & \method{} & $\Delta$ & Plain & \method{} & $\Delta$ & Plain & \method{} & $\Delta$ \\
  \midrule
  Llama-3.1-8B & 74.05 & 89.36 & \makecell{+15.31\\[-1pt]{\scriptsize[13.48, 17.02]}} & 61.31 & 78.29 & \makecell{+16.98\\[-1pt]{\scriptsize[14.02, 19.97]}} & 82.29 & 91.07 & \makecell{+8.78\\[-1pt]{\scriptsize[7.63, 10.12]}} \\
  Mistral-7B & 73.92 & 88.90 & \makecell{+14.97\\[-1pt]{\scriptsize[13.11, 17.12]}} & 66.70 & 87.55 & \makecell{+20.84\\[-1pt]{\scriptsize[18.50, 23.54]}} & 81.39 & 90.55 & \makecell{+9.16\\[-1pt]{\scriptsize[7.91, 10.40]}} \\
  \bottomrule
\end{tabular}
\caption{Argument AUROC (in points) of the two judges outside the Qwen family, with Plain input and with \method{}. $\Delta$ is \method{} minus Plain, with its 95\% CI below.}
\label{tab:other-judges}
\end{table*}

\paragraph{Deriver prompt.}
When a step calls the same function more than once, the deriver names each value it expects, such as the two hotel IDs that a request for the two cheapest hotels asks for.
The deriver gets the system prompt ``You are a careful evaluator of tool calls made by an AI assistant.''
Its user message gives the request, every earlier call and return uncut under the header ``History (all earlier calls and their returns, oldest first):'', and the function and parameter lines of the call under judgment.
It ends with this instruction:
``Before seeing any proposed value: what should the value of parameter \{name\} be in this call, and where in the context does it come from? Quote the source text if there is one, or say that the context does not determine it. Answer in at most 3 sentences.''
The message contains no value of the call under judgment and no other candidate at that step.

\paragraph{Snippet rule.}
For a candidate value, we search the full history for every place where it occurs and cut a snippet of 150 characters on each side of each occurrence; nearby windows are merged, a merged snippet is cut to 450 characters, and at most three snippets are kept, most recent first.
The test of Section~\ref{sec:follow} adds these snippets to the judge input on CFB.

\paragraph{ProvenanceGuard evidence.}
ProvenanceGuard checks a proposed call in stages, and its provenance stage asks an LLM to find, for each derivable parameter value, the evidence in the user query or earlier tool outputs that supports it \citep{she2026safeguarding}.
ProvenanceGuard then labels the call misaligned when no evidence is found.
We keep the provenance stage and hand its evidence to the judge, so that the judge returns a score.
Qwen3.8-27B sees the proposed value and quotes up to two passages that the value should come from.
It runs with thinking on, and we take the quotes from the text after the thinking.
A quote counts when it matches a piece of the full history, exactly or by a long common substring.
On FinTrace and FCRB, the snippets around the quotes follow the snippet rule above.
On CFB, it uses windows of 100 characters on each side without a length cap.
An argument without a snippet keeps its Plain score.

\paragraph{Revised calls.}

RevisEval revises the proposed call from the full history, one revised call per proposed call, and the judge reads the revised call \citep{zhang2025reviseval}.

\paragraph{The deriver as the judge.}
A natural alternative is to let the deriver judge on its own.
Table~\ref{tab:deriver-judge} lets Qwen3.8-27B judge directly, with thinking on and the same thinking budget as the deriver, over the truncated input and over the full history.
On every benchmark it stays below the same model reading the \method{} reference with thinking off, by 3.60 points on FCRB, 10.92 on CFB, and 3.58 on FinTrace with the truncated input, and by 5.74 and 2.82 with the full history.
Thinking does not close the gap either: with thinking on, the out-of-view gap of the direct judge is 12.96 on FCRB, 28.98 on CFB, and 23.17 on FinTrace.
Table~\ref{tab:grid-main} gives the scores of the other three judges reading the reference, which we compare with this direct judge.
Against the direct judge on the truncated input, Jev reading the reference is higher on all three benchmarks, by 3.5 to 4.6 points, CLM is within 1.4 points on all three, and Qwen3-8B is within 1.1 points on FCRB and CFB and 3.6 points higher on FinTrace.
Against the direct judge on the full history, Jev is 1.5 points lower on CFB and 3.4 higher on FinTrace, CLM is 4.7 and 0.6 points lower, and Qwen3-8B is 7.1 points lower on CFB and 2.6 higher on FinTrace.

\begin{table*}[!t]
\centering
\footnotesize
\setlength{\tabcolsep}{3pt}
\begin{tabular}{lccc}
  \toprule
  Qwen3.8-27B as & FCRB & CFB & FinTrace \\
  \midrule
  Direct judge, thinking on, truncated input & 89.79 & 84.17 & 88.67 \\
  Direct judge, thinking on, full history & -- & 90.27 & 89.70 \\
  Judge of the \method{} reference, thinking off & 93.39 & 95.09 & 92.25 \\
  \quad with the full history & -- & 96.01 & 92.52 \\
  \midrule
  Direct $-$ \method{}, truncated input & \makecell{$-$3.60\\[-1pt]{\scriptsize[$-$4.97, $-$2.35]}} & \makecell{$-$10.92\\[-1pt]{\scriptsize[$-$13.12, $-$8.76]}} & \makecell{$-$3.58\\[-1pt]{\scriptsize[$-$4.37, $-$2.83]}} \\
  Direct $-$ \method{}, full history & -- & \makecell{$-$5.74\\[-1pt]{\scriptsize[$-$7.40, $-$4.26]}} & \makecell{$-$2.82\\[-1pt]{\scriptsize[$-$3.57, $-$2.09]}} \\
  \bottomrule
\end{tabular}
\caption{Argument AUROC (in points) of Qwen3.8-27B judging directly with thinking on, against the same model reading the \method{} reference with thinking off. FCRB has no cut, so the full history equals the truncated input there. $\Delta$ rows give the paired difference with its 95\% CI below.}
\label{tab:deriver-judge}
\end{table*}

\begin{table}[!htbp]
  \centering
  \scriptsize
  \setlength{\tabcolsep}{2pt}
  \renewcommand{\arraystretch}{0.9}
  \begin{tabular}{lccc}
    \toprule
    Judge and evidence & FCRB & CFB & FinTrace \\
    \midrule
    \multicolumn{4}{l}{\textit{CLM}} \\
    Plain & 72.51 & 63.41 & 83.36 \\
    \makecell[l]{ProvenanceGuard evidence\\[-1pt]{\scriptsize\citep{she2026safeguarding}}} & 75.17 & 69.62 & 84.75 \\
    \makecell[l]{RevisEval\\[-1pt]{\scriptsize\citep{zhang2025reviseval}}} & 83.26 & 81.78 & 86.07 \\
    \method{} (ours) & 89.59 & 85.55 & 89.10 \\
    \method{}-v (ours) & 86.18 & 88.93 & 85.88 \\
    $\Delta$ vs.\ Plain & \makecell{+17.08\\[-1pt]{\scriptsize[15.06, 19.22]}} & \makecell{+22.14\\[-1pt]{\scriptsize[19.75, 24.74]}} & \makecell{+5.74\\[-1pt]{\scriptsize[4.08, 7.54]}} \\
    $\Delta$ vs.\ RevisEval & \makecell{+6.33\\[-1pt]{\scriptsize[4.65, 7.98]}} & \makecell{+3.77\\[-1pt]{\scriptsize[1.72, 5.90]}} & \makecell{+3.03\\[-1pt]{\scriptsize[1.42, 4.79]}} \\
    \midrule
    \multicolumn{4}{l}{\textit{Jev}} \\
    Plain & 90.83 & 75.01 & 89.11 \\
    Full history & -- & 79.82 & 90.08 \\
    ProvenanceGuard evidence & 91.00 & 80.74 & 89.57 \\
    RevisEval & 92.23 & 87.15 & 91.77 \\
    \method{} (ours) & 93.30 & 88.73 & 93.10 \\
    \method{}-v (ours) & 91.81 & 92.23 & 93.54 \\
    $\Delta$ vs.\ Plain & \makecell{+2.47\\[-1pt]{\scriptsize[1.55, 3.39]}} & \makecell{+13.72\\[-1pt]{\scriptsize[11.56, 15.79]}} & \makecell{+3.99\\[-1pt]{\scriptsize[3.23, 4.85]}} \\
    $\Delta$ vs.\ RevisEval & \makecell{+1.07\\[-1pt]{\scriptsize[$-$0.05, 2.13]}} & \makecell{+1.57\\[-1pt]{\scriptsize[0.31, 2.98]}} & \makecell{+1.33\\[-1pt]{\scriptsize[0.77, 1.88]}} \\
    \midrule
    \multicolumn{4}{l}{\textit{Qwen3-8B}} \\
    Plain & 78.79 & 66.79 & 85.15 \\
    Full history & -- & 66.92 & 84.74 \\
    \makecell[l]{Self-generated reference\\[-1pt]{\scriptsize\citep{zeng2024llmbar}}} & 81.80 & 68.28 & 87.95 \\
    ProvenanceGuard evidence & 77.71 & 64.80 & 85.21 \\
    RevisEval & 87.06 & 81.75 & 90.52 \\
    \method{} (ours) & 89.48 & 83.13 & 92.28 \\
    \method{}-v (ours) & 85.99 & 87.89 & 91.72 \\
    $\Delta$ vs.\ Plain & \makecell{+10.69\\[-1pt]{\scriptsize[8.85, 12.55]}} & \makecell{+16.34\\[-1pt]{\scriptsize[13.83, 18.69]}} & \makecell{+7.13\\[-1pt]{\scriptsize[6.00, 8.29]}} \\
    $\Delta$ vs.\ RevisEval & \makecell{+2.42\\[-1pt]{\scriptsize[0.89, 4.01]}} & \makecell{+1.38\\[-1pt]{\scriptsize[$-$0.64, 3.26]}} & \makecell{+1.75\\[-1pt]{\scriptsize[0.97, 2.53]}} \\
    \midrule
    \multicolumn{4}{l}{\textit{Qwen3.8-27B}} \\
    Plain & 91.95 & 78.92 & 89.31 \\
    Full history & -- & 85.45 & 89.77 \\
    Self-generated reference & 92.52 & 81.83 & 90.05 \\
    ProvenanceGuard evidence & 91.94 & 88.08 & 90.09 \\
    RevisEval & 93.70 & 93.54 & 92.53 \\
    \method{} (ours) & 93.39 & 95.09 & 92.25 \\
    \method{}-v (ours) & 91.50 & 94.18 & 92.22 \\
    $\Delta$ vs.\ Plain & \makecell{+1.44\\[-1pt]{\scriptsize[0.37, 2.51]}} & \makecell{+16.17\\[-1pt]{\scriptsize[13.89, 18.31]}} & \makecell{+2.94\\[-1pt]{\scriptsize[2.04, 3.88]}} \\
    $\Delta$ vs.\ RevisEval & \makecell{$-$0.32\\[-1pt]{\scriptsize[$-$1.48, 0.80]}} & \makecell{+1.55\\[-1pt]{\scriptsize[0.25, 2.83]}} & \makecell{$-$0.27\\[-1pt]{\scriptsize[$-$0.93, 0.38]}} \\
    \bottomrule
  \end{tabular}
  \caption{Argument AUROC (in points) of four judges with each kind of evidence. ProvenanceGuard evidence and RevisEval are the prior methods (Section~\ref{sec:setup}). \method{}-v lets the deriver see the candidate value (Section~\ref{sec:paths}). Full history gives Jev and the LLM judges the uncut history (Appendix~\ref{app:data}); FCRB has no cut, so there it equals Plain. Self-generated reference has the judge write its own reference from its own input (Section~\ref{sec:setup}). $\Delta$ rows give the difference of \method{} against Plain and against RevisEval, the stronger prior method in every cell, with its 95\% CI below. Table~\ref{tab:main} gives the main inputs, and Table~\ref{tab:grid-call} gives call AUROC.}
  \label{tab:grid-main}
\end{table}
\begin{table}[!htbp]
  \centering
  \footnotesize
  \begin{tabular}{lcccc}
    \toprule
    Judge & Plain & Full history & \method{} & $\Delta$ \\
    \midrule
    \multicolumn{5}{l}{\textit{FCRB}, \textsc{Default} (185 correct)} \\
    CLM & 31.93 & -- & 3.80 & \makecell{$-$28.13\\[-1pt]{\scriptsize[$-$36.67, $-$20.70]}} \\
    Jev & 20.26 & -- & 10.28 & \makecell{$-$9.98\\[-1pt]{\scriptsize[$-$15.63, $-$4.20]}} \\
    Qwen3-8B & 14.40 & -- & $-$3.93 & \makecell{$-$18.33\\[-1pt]{\scriptsize[$-$25.50, $-$9.25]}} \\
    Qwen3.8-27B & 36.03 & -- & 12.12 & \makecell{$-$23.91\\[-1pt]{\scriptsize[$-$32.30, $-$15.69]}} \\
    \midrule
    \multicolumn{5}{l}{\textit{CFB}, \textsc{History-only} (279 correct)} \\
    CLM & 66.63 & -- & 28.26 & \makecell{$-$38.37\\[-1pt]{\scriptsize[$-$47.49, $-$30.14]}} \\
    Jev & 59.39 & \makecell{21.89\\[-1pt]{\scriptsize[13.69, 29.20]}} & 38.35 & \makecell{$-$21.04\\[-1pt]{\scriptsize[$-$30.54, $-$13.38]}} \\
    Qwen3-8B & 74.87 & \makecell{12.03\\[-1pt]{\scriptsize[6.66, 18.78]}} & 31.37 & \makecell{$-$43.50\\[-1pt]{\scriptsize[$-$55.21, $-$33.14]}} \\
    Qwen3.8-27B & 57.08 & \makecell{3.36\\[-1pt]{\scriptsize[$-$2.76, 9.20]}} & 26.21 & \makecell{$-$30.87\\[-1pt]{\scriptsize[$-$40.51, $-$21.81]}} \\
    \midrule
    \multicolumn{5}{l}{\textit{FinTrace}, \textsc{No-source} (2,150 correct)} \\
    CLM & 28.28 & -- & 30.68 & \makecell{+2.41\\[-1pt]{\scriptsize[$-$0.43, 5.55]}} \\
    Jev & 27.47 & \makecell{29.09\\[-1pt]{\scriptsize[26.32, 31.64]}} & 27.68 & \makecell{+0.21\\[-1pt]{\scriptsize[$-$1.91, 2.18]}} \\
    Qwen3-8B & 25.72 & \makecell{29.04\\[-1pt]{\scriptsize[26.70, 31.54]}} & 26.41 & \makecell{+0.69\\[-1pt]{\scriptsize[$-$2.18, 3.12]}} \\
    Qwen3.8-27B & 31.68 & \makecell{34.23\\[-1pt]{\scriptsize[31.56, 36.81]}} & 29.13 & \makecell{$-$2.55\\[-1pt]{\scriptsize[$-$4.86, $-$0.12]}} \\
    \bottomrule
  \end{tabular}
  \caption{Derivation gap $\mathrm{Gap}_\tau$ (in points) on the main path of each benchmark, with $\tau$ set for each input. Full history gives Jev and the LLM judges the uncut history (Appendix~\ref{app:data}), with its 95\% CI below; CLM cannot read it, and FCRB has no cut. $\Delta$ is \method{} minus Plain, with its 95\% CI below. On FCRB, the acceptance at $\tau$ of the 185 correct defaults is 52.43, 64.08, 65.95, and 50.54 with Plain input and 78.92, 76.03, 86.20, and 73.35 with \method{}, for the four judges in table order; for the 793 correct values in view it is 84.36, 84.34, 80.35, and 86.57 with Plain input and 82.72, 86.31, 82.27, and 85.47 with \method{}.}
  \label{tab:grid-gap}
\end{table}
\begin{table*}[!htbp]
\centering
\footnotesize
\begin{tabular}{lcccc}
  \toprule
  Input & CLM & Jev & Qwen3-8B & Qwen3.8-27B \\
  \midrule
  \multicolumn{5}{l}{\textit{CFB}, \textsc{History-only} (279 correct, 59 wrong)} \\
  Plain (AUROC) & 64.33 & 67.78 & 49.18 & 77.56 \\
  ProvenanceGuard evidence $-$ Plain & \makecell{+5.01\\[-1pt]{\scriptsize[$-$4.04, 13.34]}} & \makecell{+7.81\\[-1pt]{\scriptsize[0.40, 16.07]}} & \makecell{+17.64\\[-1pt]{\scriptsize[7.83, 26.98]}} & \makecell{+4.25\\[-1pt]{\scriptsize[$-$2.44, 11.69]}} \\
  RevisEval $-$ Plain & \makecell{+23.93\\[-1pt]{\scriptsize[16.90, 30.81]}} & \makecell{+26.18\\[-1pt]{\scriptsize[17.28, 36.16]}} & \makecell{+37.45\\[-1pt]{\scriptsize[27.23, 47.61]}} & \makecell{+18.33\\[-1pt]{\scriptsize[10.81, 25.64]}} \\
  Full history $-$ Plain & -- & \makecell{+12.80\\[-1pt]{\scriptsize[2.93, 23.17]}} & \makecell{+25.95\\[-1pt]{\scriptsize[16.29, 34.90]}} & \makecell{+11.45\\[-1pt]{\scriptsize[3.90, 19.93]}} \\
  \method{} $-$ Plain & \makecell{+27.16\\[-1pt]{\scriptsize[17.63, 36.74]}} & \makecell{+21.83\\[-1pt]{\scriptsize[13.32, 30.54]}} & \makecell{+43.69\\[-1pt]{\scriptsize[34.07, 53.29]}} & \makecell{+16.87\\[-1pt]{\scriptsize[9.93, 23.04]}} \\
  \midrule
  \multicolumn{5}{l}{\textit{CFB}, \textsc{No-source} (462 correct, 747 wrong)} \\
  Plain (AUROC) & 58.13 & 77.14 & 61.40 & 75.46 \\
  ProvenanceGuard evidence $-$ Plain & \makecell{+5.49\\[-1pt]{\scriptsize[1.32, 9.83]}} & \makecell{+2.23\\[-1pt]{\scriptsize[0.51, 4.04]}} & \makecell{$-$13.18\\[-1pt]{\scriptsize[$-$16.95, $-$9.43]}} & \makecell{+5.36\\[-1pt]{\scriptsize[3.93, 6.97]}} \\
  RevisEval $-$ Plain & \makecell{+19.87\\[-1pt]{\scriptsize[15.49, 24.39]}} & \makecell{+13.23\\[-1pt]{\scriptsize[10.55, 16.08]}} & \makecell{+18.75\\[-1pt]{\scriptsize[15.10, 22.67]}} & \makecell{+16.93\\[-1pt]{\scriptsize[13.50, 20.62]}} \\
  Full history $-$ Plain & -- & \makecell{$-$2.83\\[-1pt]{\scriptsize[$-$4.51, $-$1.28]}} & \makecell{$-$13.83\\[-1pt]{\scriptsize[$-$18.90, $-$9.00]}} & \makecell{$-$0.28\\[-1pt]{\scriptsize[$-$2.56, 2.05]}} \\
  \method{} $-$ Plain & \makecell{+25.04\\[-1pt]{\scriptsize[20.15, 29.88]}} & \makecell{+11.10\\[-1pt]{\scriptsize[8.38, 13.84]}} & \makecell{+18.82\\[-1pt]{\scriptsize[14.49, 23.44]}} & \makecell{+17.80\\[-1pt]{\scriptsize[14.00, 21.82]}} \\
  \bottomrule
\end{tabular}
\caption{Argument AUROC (in points) on the two main paths of CFB. The first row of each block gives AUROC with Plain input, and the other rows give differences with their 95\% CIs below. The wrong values of a path are those whose own value lies on that path. An argument with no item for an input keeps its Plain score. -- marks inputs that we do not run with that judge.}
\label{tab:grid-cfb}
\end{table*}

\paragraph{Self-generated reference.}
Following LLMBar \citep{zeng2024llmbar}, the judge model first writes the expected value from its own truncated input, with thinking off, greedy decoding, and at most 384 new tokens.
It then reads that reference as it would read \method{}.

\section{Full Comparison Grid}
\label{app:grid}
Table~\ref{tab:grid-main} gives argument AUROC for each pair of judge and benchmark with every input, extending Table~\ref{tab:main}.
Averaged over the four judges, \method{} minus Plain is 7.92 [6.72, 9.07] on FCRB, 17.09 [15.47, 18.85] on CFB, and 4.95 [4.00, 6.02] on FinTrace; minus ProvenanceGuard evidence it is 7.48 [6.28, 8.63], 12.31 [10.88, 13.81], and 4.28 [3.35, 5.27]; and minus RevisEval it is 2.38 [1.29, 3.48], 2.07 [0.71, 3.47], and 1.46 [0.67, 2.27].
Averaged over the three judges that read the full history, \method{} minus the full history is 11.58 [10.22, 12.97] on CFB and 4.35 [3.55, 5.22] on FinTrace.
ProvenanceGuard evidence and RevisEval are the prior methods of Section~\ref{sec:setup}.
We run both with all four judges, and the full history with every judge but CLM.
In Table~\ref{tab:grid-main}, RevisEval has the higher AUROC of the two prior methods for every judge and benchmark, so $\Delta$ vs.\ RevisEval is also the difference against the best prior method.
On CFB, \method{} minus the full history is 8.91 [7.27, 10.62] for Jev, 16.20 [14.03, 18.30] for Qwen3-8B, and 9.64 [7.81, 11.38] for Qwen3.8-27B.
The self-generated reference minus Plain is 3.01 [1.59, 4.52], 1.49 [$-$0.06, 3.20], and 2.80 [2.11, 3.57] for Qwen3-8B on FCRB, CFB, and FinTrace, and 0.57 [$-$0.05, 1.18], 2.92 [1.76, 4.11], and 0.74 [0.15, 1.30] for Qwen3.8-27B.
\method{} minus the self-generated reference is 7.68 [5.92, 9.40], 14.85 [11.91, 17.62], and 4.33 [3.52, 5.14] for Qwen3-8B, and 0.87 [$-$0.08, 1.84], 13.26 [10.99, 15.53], and 2.20 [1.24, 3.25] for Qwen3.8-27B.
\method{} minus ProvenanceGuard evidence on CFB is 15.93 [14.15, 18.02] for CLM, 7.99 [6.16, 9.76] for Jev, 18.33 [16.15, 20.29] for Qwen3-8B, and 7.01 [5.09, 8.88] for Qwen3.8-27B.
Every CI of \method{} against ProvenanceGuard evidence, on all three benchmarks, lies above zero.
Table~\ref{tab:grid-gap} gives the derivation gap on the main path of each benchmark, and Table~\ref{tab:grid-cfb} gives each kind of evidence on the two main paths of CFB, extending the ablation of Section~\ref{sec:paths}.

\clearpage
\begin{figure*}[!htbp]
\centering
\begin{minipage}{0.97\textwidth}
\footnotesize
\textbf{Judge input (Plain), excerpt}\par\vspace{1pt}
\begin{Verbatim}[breaklines=true, breakanywhere=true, fontsize=\scriptsize, breaksymbolleft={}, frame=single, framesep=2pt]
Question: Could you provide me with the details of Adriel's internship experiences, including the name of the organizations and the dates of the internships?
Function: get_adriel_detail_experience_and_education -- Retrieves detailed information about Adriel's professional experiences and educational background, including the type of experience or education, name of the organization or institution, and relevant dates.
Parameter: experience_or_education_name -- The name of the organization where the experience was gained or the institution where the education was obtained.; type: string; default: "Not specified"
Proposed value for experience_or_education_name: "Not specified"
\end{Verbatim}
\textbf{Line added for schema facts}\par\vspace{1pt}
\begin{Verbatim}[breaklines=true, breakanywhere=true, fontsize=\scriptsize, breaksymbolleft={}, frame=single, framesep=2pt]
Schema default for experience_or_education_name: "Not specified". The question does not contain any of these words: "experience_or_education_name", "experience or education name", "Not specified".
\end{Verbatim}
\textbf{\method{} reference}\par\vspace{1pt}
\begin{Verbatim}[breaklines=true, breakanywhere=true, fontsize=\scriptsize, breaksymbolleft={}, frame=single, framesep=2pt]
The question asks broadly for all of Adriel's internship experiences without naming any specific organization, so the context does not determine a particular value for `experience_or_education_name`. The parameter should take its default value of "Not specified" to retrieve all matching internship details.
\end{Verbatim}
\vspace{2pt}
\centering
\begin{tabular}{lccc}
  \toprule
  Judge & Plain & Schema facts & \method{} \\
  \midrule
    CLM & 0.3718 / 0.6567 & 0.5092 / 0.6640 & 0.8924 / 0.7014 \\
    Jev & 0.7300 / 0.8600 & 0.7300 / 0.8600 & 0.9600 / 0.9300 \\
    Qwen3-8B & 0.0000 / 1.0000 & 0.0032 / 1.0000 & 1.0000 / 1.0000 \\
    Qwen3.8-27B & 0.1824 / 0.6514 & 0.1824 / 0.4688 & 0.9903 / 0.9627 \\
  \bottomrule
\end{tabular}
\end{minipage}
\caption{A correct default on FCRB (\texttt{live\_multiple\_171-71-0}). Each cell gives the judge's score and its threshold $\tau$ for that input (score / $\tau$). With \method{}, the score of Qwen3-8B equals its $\tau$.}
\label{fig:case-default}
\end{figure*}

\section{Extended Related Work}
\label{app:related}
This appendix expands the related work of Section~\ref{sec:related}.
\paragraph{Judges for tool-calling.}
Tool-call benchmarks check calls against gold answers.
BFCL matches each argument value against a set of allowed answers \citep{patil2025berkeley}, and ComplexFuncBench judges a whole call with rules, response matching, and an LLM \citep{zhong2025complexfuncbench}.
Without gold answers, learned judges score or compare whole calls \citep{agarwal2025toolrm,li2026toolrm} or score each argument as a step \citep{lin2026toolprm}.
ToolPRMBench and AgentProcessBench evaluate step-level judges for tool-using agents, and the latter finds that judges tend to label steps as correct \citep{li2026toolprmbench,fan2026agentprocessbench}.
AgentJudgeBench finds that LLM judges of tool calls agree less with a programmatic reference as tasks get harder \citep{verma2026agentjudgebench}.
Guards check calls before they run.
ProvenanceGuard has an LLM find evidence for each derivable argument with the value in hand, and later, without the proposed action, check whether more than one action fits the context \citep{she2026safeguarding}.
PACT tracks where each argument value came from \citep{fan2026granularity}.
In Agent-Sentry, the agent reports whether each value is copied verbatim, derived, or generated \citep{sequeira2026agentsentry}.
When rules flag a suspected failure in a proposed call, TwinCheck repairs it or writes a constrained alternative from the trace and the flagged evidence, and a verifier compares the two \citep{dai2026twincheck}.
Jev trails GPT-6 where the verdict has to be derived rather than read off the text \citep{li2026jevjudge}, and factuality metrics drop on summaries that need deeper reasoning to check \citep{ramprasad2025automatic}.
To our knowledge, none of these studies splits judge errors by derivation path.

\paragraph{Reference-guided judging.}
Judges grade better with a reference answer.
In MT-Bench, the judge first answers a math question without seeing the answers under review, then grades against its own answer, which reduces its errors \citep{zheng2023judging}.
LLMBar and the candidate-blind head of BAER also have the judge write its own output first \citep{zeng2024llmbar,li2026baer}, and RevisEval builds the reference by revising the candidate \citep{zhang2025reviseval}.
Without a reference, judges agree less with experts \citep{krumdick2025free} and give credit to wrong answers \citep{chalamalasetti2026llm}.
On math and code, judges without a reference reward wrong answers that look convincing, and a judge that commits to its own solution before grading accepts far fewer of them \citep{zhou2026convincing}.
SGV finds that verifiers of agent trajectories tend to agree with the candidate, and it first writes priors without the candidate \citep{andrade2026let}.
MARCH and factored Chain-of-Verification also check claims without showing the checker the answer under test \citep{li2026march,dhuliawala2024chain}.
For code agents, an agent explores the repository and writes a rubric before scoring any patch \citep{raghavendra2026agentic}.
When a QA reference is swapped for a wrong one, judges often reject answers that match it, which the authors trace to the judges' own knowledge \citep{lee2026judging}.
These judges write the reference from the question and their own knowledge, which is often not enough for an argument judge, whose expected value can sit in a tool return cut from its input.
A reference that the judge writes from its own input, as in LLMBar, improves on Plain input by at most 3.01 points but stays below \method{} (Section~\ref{sec:main}); writing the \method{} reference from the cut input instead of the full history alone costs 9.63 points on CFB (Table~\ref{tab:ablation}), and on CFB and FCRB our judges follow a swapped reference (Section~\ref{sec:swap}).

\paragraph{Correct content without support.}
A value without support in the context is not always wrong.
Many entities that summarizers add without support are still factual \citep{cao2022hallucinated}.
RAG models also often answer correctly when the context is insufficient \citep{joren2025sufficient}.
Models also score an inconsistent summary higher when its text is copied from the source \citep{tam2023evaluating}.
Tool-call arguments make this common, since a correct default, format change, or inferred value has no matching span in the dialogue.
We do not treat such values as errors, and we measure how often judges do.
A source in the context can also go unused.
Models miss facts that share few words with the question \citep{modarressi2025nolima}, and they do worse as tool returns grow longer \citep{kate2025longfunceval}.

\section{Qualitative Case Studies}
\label{app:cases}
Figures~\ref{fig:case-default} and~\ref{fig:case-swap} show two arguments, one for each of two findings.
Figure~\ref{fig:overview} already shows a correct value whose source was cut.
For each finding, we wrote a condition before reading any score and drew one argument at random from those that meet it, with seed 20260930.
A script copies all text from the judge inputs and the deriver outputs.
The excerpts show the lines relevant to the argument, and as in Figure~\ref{fig:input} we write the long dash after a function or parameter name as \texttt{-{}-}.
Scores are each judge's probability that the value is correct.

\paragraph{A default.}
In Figure~\ref{fig:case-default}, the request asks for the details of Adriel's internships and names no organization.
The correct value of \texttt{experience\_or\_education\_name} is the schema default, ``Not specified'', which the parameter line states.
With Plain input and with the schema-facts line, all four judges score it below their thresholds $\tau$.
The \method{} reference says that the request names no organization, so the parameter should take its default.
With it, CLM, Jev, and Qwen3.8-27B score the value above their thresholds, and Qwen3-8B scores it at its threshold of 1.0000, which counts as accepted only with the tie fraction $q$ (Appendix~\ref{app:metrics}).
The condition is this pattern for all four judges.
One of the 142 arguments of Figure~\ref{fig:follow}b meets it.

\paragraph{A swapped reference.}
In Figure~\ref{fig:case-swap}, the request asks for an alarm at 5 in the afternoon.
The correct value of \texttt{new\_alarm\_time} is 17:00, and the wrong FCRB call, made by a real model, uses 15:00.
The original reference names 17:00, and the swapped reference changes only that value.
With the original reference, all four judges score the correct value at or above 0.5 and the wrong value below 0.5.
With the swapped reference, all four judges reverse both judgments.
Under Plain input, CLM and Qwen3-8B score both values at or above 0.5, Jev scores only the correct value above 0.5, and Qwen3.8-27B scores only the wrong value above 0.5.
The condition is that all four judges reverse both judgments at 0.5.
Of the 749 CFB and FCRB arguments in the swap test, 14 meet it.
We exclude the two pairs whose values differ only by a power of ten, 5.0 against 0.05 and 16 against 1600, since the schema leaves their unit open, and draw from the remaining 12.

\begin{figure*}[!htbp]
\centering
\begin{minipage}{0.97\textwidth}
\footnotesize
\textbf{Judge input for the correct value (Plain), excerpt}\par\vspace{1pt}
\begin{Verbatim}[breaklines=true, breakanywhere=true, fontsize=\scriptsize, breaksymbolleft={}, frame=single, framesep=2pt]
Question: Could you set a new alarm for 5 in the afternoon labeled 'Grocery run'?
Previous calls (most recent 5):
none
Function: Alarm_1_AddAlarm -- This function sets a new alarm with a specified time and an optional name.
Parameter: new_alarm_time -- The time to set for the new alarm, in 24-hour format (HH:MM).; type: string
Proposed value for new_alarm_time: "17:00"
\end{Verbatim}
\textbf{Original \method{} reference}\par\vspace{1pt}
\begin{Verbatim}[breaklines=true, breakanywhere=true, fontsize=\scriptsize, breaksymbolleft={}, frame=single, framesep=2pt]
The value should be **"17:00"**, since the parameter requires 24-hour HH:MM format. This is derived from the user's request text: "set a new alarm for **5 in the afternoon**."
\end{Verbatim}
\textbf{Swapped reference}\par\vspace{1pt}
\begin{Verbatim}[breaklines=true, breakanywhere=true, fontsize=\scriptsize, breaksymbolleft={}, frame=single, framesep=2pt]
The value should be **"15:00"**, since the parameter requires 24-hour HH:MM format. This is derived from the user's request text: "set a new alarm for **5 in the afternoon**."
\end{Verbatim}
\vspace{2pt}
\centering
\begin{tabular}{lcccccc}
  \toprule
  & \multicolumn{2}{c}{Plain} & \multicolumn{2}{c}{Original reference} & \multicolumn{2}{c}{Swapped reference} \\
  \cmidrule(lr){2-3}\cmidrule(lr){4-5}\cmidrule(lr){6-7}
  Judge & Correct & Wrong & Correct & Wrong & Correct & Wrong \\
  \midrule
    CLM & 0.7870 & 0.7632 & 0.8059 & 0.4389 & 0.3369 & 0.8281 \\
    Jev & 0.9800 & 0.0600 & 0.9800 & 0.0100 & 0.4100 & 0.9700 \\
    Qwen3-8B & 1.0000 & 1.0000 & 1.0000 & 0.0000 & 0.0000 & 1.0000 \\
    Qwen3.8-27B & 0.3486 & 0.9579 & 0.9914 & 0.0003 & 0.0005 & 0.9890 \\
  \bottomrule
\end{tabular}
\end{minipage}
\caption{A swapped reference on FCRB (\texttt{live\_multiple\_809-176-0}). The correct value is 17:00 and the wrong value is 15:00. Cells give scores.}
\label{fig:case-swap}
\end{figure*}

\end{document}